\documentclass[journal]{IEEEtran}
\usepackage{amsmath,amsfonts}
\usepackage{algorithmic}
\usepackage{algorithm}
\usepackage{array}
\usepackage[caption=false,font=normalsize,labelfont=sf,textfont=sf]{subfig}
\usepackage{textcomp}
\usepackage{multirow}
\usepackage{stfloats}
\usepackage{url}
\usepackage{verbatim}
\usepackage{graphicx}
\usepackage{cite}
\usepackage{colortbl}
\usepackage{color}
\usepackage{xcolor}
\usepackage{makecell}
\usepackage{booktabs}
\usepackage{arydshln}
\usepackage[breaklinks,colorlinks]{hyperref}
\usepackage{soul}
\newcommand{\PreserveBackslash}[1]{\let\temp=\\#1\let\\=\temp}

\newcolumntype{C}[1]{>{\PreserveBackslash\centering}p{#1}}
\newcolumntype{R}[1]{>{\PreserveBackslash\raggedleft}p{#1}}
\newcolumntype{L}[1]{>{\PreserveBackslash\raggedright}p{#1}}

\begin{document}
%
\title{DPF-Net: Physical Imaging Model Embedded Data-Driven Underwater Image Enhancement}
%
%
%

\author{Han~Mei,
        Kunqian~Li,~\IEEEmembership{Member,~IEEE},
        Shuaixin~Liu,
        Chengzhi~Ma,
        Qianli~Jiang 
\thanks{The research has been supported by the National Natural Science Foundation of China under Grant 62371431 and 61906177, in part by the Marine Industry Key Technology Research and Industrialization Demonstration Project of Qingdao under Grant 23-1-3-hygg-20-hy, and in part by the Fundamental Research Funds for the Central Universities under Grants 202262004. (Corresponding author: Kunqian Li and Qianli Jiang)}

\thanks{Han Mei, Kunqian Li, Shuaixin Liu and Chengzhi~Ma are with the College of Engineering, Ocean University of China, Qingdao 266404, China (meihan@stu.ouc.edu.cn; likunqian@ouc.edu.cn; liushuaixin@stu.ouc.edu.cn; machengzhi@stu.ouc.edu.cn).}
\thanks{Qianli Jiang is with Institute for Advanced Ocean Study, Ocean University of China, Qingdao 266100, China (jiangqianli@ouc.edu.cn).}

}
%
%

\markboth{ }%
{Shell \MakeLowercase{\textit{et al.}}: Bare Dem of IEEEtran.cls for Journals}
%



\maketitle

\begin{abstract}
Due to the complex interplay of light absorption and scattering in the underwater environment, underwater images experience significant degradation. This research presents a two-stage underwater image enhancement network called the Data-Driven and Physical Parameters Fusion Network (DPF-Net), which harnesses the robustness of physical imaging models alongside the generality and efficiency of data-driven methods. We first train a physical parameter estimate module using synthetic datasets to guarantee the trustworthiness of the physical parameters, rather than solely learning the fitting relationship between raw and reference images by the application of the imaging equation, as is common in prior studies. This module is subsequently trained in conjunction with an enhancement network, where the estimated physical parameters are integrated into a data-driven model within the embedding space. To maintain the uniformity of the restoration process amid underwater imaging degradation, we propose a physics-based degradation consistency loss. Additionally, we suggest an innovative weak reference loss term utilizing the entire dataset, which alleviates our model's reliance on the quality of individual reference images. Our proposed DPF-Net demonstrates superior performance compared to other benchmark methods across multiple test sets, achieving state-of-the-art results. The source code and pre-trained models are available on the project home page: \url{https://github.com/OUCVisionGroup/DPF-Net}.
\end{abstract}


%
\section{Introduction}
\IEEEPARstart{A}{s} marine resources are continuously developed and utilized, underwater imagery is increasingly employed in marine military, environmental protection, engineering, and other domains. Nonetheless, the unique characteristics of the underwater environment, including water refraction, scattering, and absorption, result in subpar image quality, manifesting as low contrast, color distortion, and detail blurring \cite{jian2021underwater, WANG2024Self-organized}. These issues not only impair human visual perception of the underwater world but also hinder the successful advancement of underwater operations, resource exploration, ecological conservation, and other activities. Thus, deriving significant information from such compromised underwater images is an exceedingly challenging endeavor. Enhancing the quality of underwater images and improving their visual impact have emerged as critical issues that necessitate immediate attention and resolution.

Traditional Underwater Image Enhancement (UIE) methods primarily include contrast enhancement, noise reduction, and sharpening, among other techniques that directly modify pixel values \cite{hitam2013mixture, li2015underwater, WANG2023Meta, CHANG2023UIDEF}. These procedures can improve the visual quality of the image to some extent. However, due to the complex and variable underwater imaging conditions, such as illumination, water quality, and other factors, these technologies are highly sensitive and may not consistently achieve the desired outcomes. Unlike terrestrial images, underwater images suffer degradation that involves both light attenuation and scattering effects. Akkaynak et al. \cite{akkaynak2018revised} proposed a physical model for underwater imaging, decomposing images into forward scattered and backscattered components, which serves as the theoretical foundation of this paper. Then, they further established a comprehensive and robust theoretical framework that leverages physical imaging knowledge to enhance underwater images, thereby producing more reliable and accurate results \cite{Akkaynak2019SeaThru}. However, for underwater images characterized by varied degrees and types of degradation, accurately and efficiently determining the numerous degraded parameters of the physical imaging model remains the primary obstacle limiting the widespread adoption of these methods.

In recent years, the rapid advancement of deep learning technology has made deep learning-based underwater image enhancement techniques a significant area of research \cite{anwar2020diving}. By training on paired underwater image sets or domains with unpaired images, these methods can learn the characteristics and underlying principles of underwater images and their relationships with clear images or domains, thereby enabling adaptive visual enhancement of the images. This data-driven enhancement strategy has demonstrated generality in various degradation scenarios. However, because the learned references are often not truly reliable clear images, the enhancements may not accurately reflect the true degradation characteristics.

To address these issues, we propose a UIE network that integrates physical parameters from the underwater imaging model, referred to as the Data-Driven and Physical Parameters Fusion Network (DPF-Net). Our primary contribution can be summarized as follows.
  \begin{itemize}
    \item [(1)] We have developed an improved fusion network for underwater image enhancement, termed DPF-Net, which integrates data-driven approaches with physical imaging models. This enables the network to learn various aspects of underwater scenes from extensive datasets while also uncovering deterioration correlations through physical parameters. Physical parameters are incorporated into the training of this advanced model via an embedding space approach. The alignment between model enhancement and physical imaging is ensured by employing a degradation consistency loss.
    \item [(2)] A Degraded Parameters Estimation Module (DPEM) is proposed, which is trained on a synthetic dataset to ensure that the predicted parameters are consistent with the physical model of underwater imaging. The module contributes to the prediction of physical parameters in DPF-Net as a pre-trained model, rather than undergoing synchronous training from scratch to acquire the degradation fit of the paired raw-reference images.
    \item [(3)] In response to the limitations of reference images in existing public UIE datasets, we have developed a suite of loss functions. This includes regionally weighted reference losses and weakly supervised reference losses derived from statistical analyses of comprehensive datasets. These innovations not only reduce the dependence on pixel-level fidelity of reference images but also enhance the vibrancy of the resulting enhanced images.
  \end{itemize}

\section{Related Works}\label{sec:Rel}
\subsection{Physical Model of Underwater Imaging}\label{sec:Rel_A}
\begin{figure}[t]
  \centering
   \includegraphics[width=0.9\linewidth]{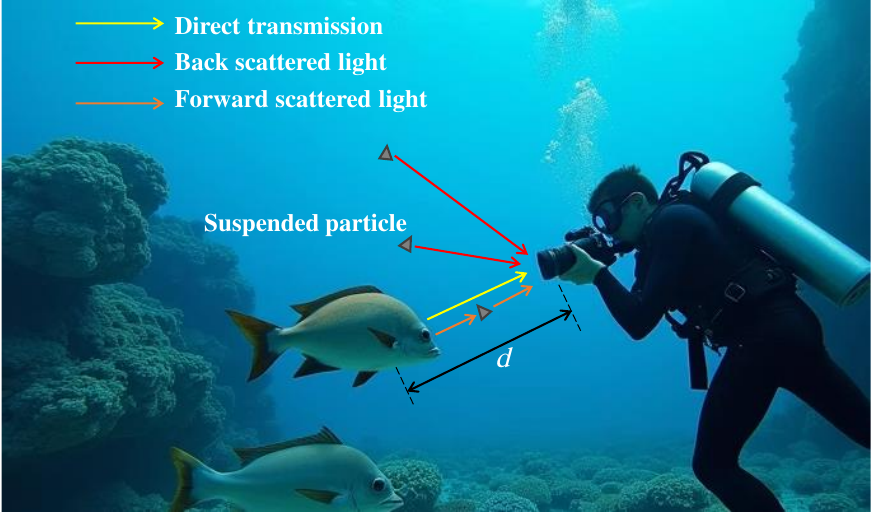}
   \caption{Schematic diagram of underwater imaging process.}
   \label{fig:imaging_process}
\end{figure}

According to \cite{akkaynak2017space}, the physical model of underwater imaging must comprehensively account for both the absorption of light by water and the scattering of light by suspended particles. The light captured by the camera typically comprises direct transmission light, backscattered light, and forward-scattered light, as illustrated in Figure \ref{fig:imaging_process}. Given that forward-scattered light has a minimal impact on image degradation, it is not considered in this analysis. The physical model for underwater imaging incorporates several critical parameters: global background light $B^{\infty}$, attenuation coefficient $\beta^D$, scattering coefficient $\beta^B$, and absolute depth $d$. This model can be formally described as follows:
\begin{equation}
    I = J \cdot e^{-\beta^{D} \cdot d} + B^{\infty} \cdot (1 - e^{-\beta^{B} \cdot d}),
    \label{eq1}
\end{equation}
where $J$ represents scene radiance (clear image) and $I$ represents radiance captured by the camera (raw image). It should be noted that there have been some previous works \cite{song2020enhancement, fu2022unsupervised, cong2023pugan} to simplify the model, assuming that $\beta^D = \beta^B$ and $e^{-\beta^{D} \cdot d} = e^{-\beta^{B} \cdot d} = T$, which is called the transmission map. As highlighted in the literature \cite{solonenko2015inherent}, $\beta^D$ and $\beta^B$ demonstrate significant disparities in both their physical interpretations and numerical values. The attenuation and scattering coefficients vary markedly across different water types, which cannot be simply equated.

In the widely adopted underwater image synthesis procedure, also utilized in this paper, the parameters $\beta^D$ and $\beta^B$ are derived from ten Jerlov water types described in \cite{solonenko2015inherent}. The parameter values for each water type are obtained from data at wavelengths corresponding to 650 nm (red), 525 nm (green), and 450 nm (blue), which align with the RGB channels, respectively. For the pre-estimation of $B^{\infty}$, following the methodology outlined in \cite{song2020enhancement}, $B^{\infty}$ can be calculated using the following formula:
\begin{equation}
    B^{\infty}_r = 140 / (1 + 14.4 \times e^{-0.034 \times Med_r}),
    \label{eq2}
\end{equation}
\begin{equation}
    B^{\infty}_{b,g} = 1.13 \times Avg_{b,g} + 1.11 \times Std_{b,g} - 25.6,
    \label{eq3}
\end{equation}
where $B^{\infty}_r$ denotes the background light in the red channel, $Med_r$ represents the median value of pixel intensities in the Red channel, and $B^{\infty}_{b,g}$ signifies the background light in the Blue or Green channels, computed independently. Additionally, $Avg_{b,g}$ and $Std_{b,g}$ denote the mean and standard deviation of the Blue or Green channels, respectively. The $B^{\infty}$ values obtained through Equations (\ref{eq2}) and (\ref{eq3}) are not precise; therefore, they are used only for an initial estimation of $B^{\infty}$. 

The physical model of underwater imaging provides a reliable theoretical reference for underwater image restoration and enhancement. However, how to effectively determine numerous degraded parameters for degraded images is the key to its effective application in the enhancement process.

\subsection{Deep Learning-based Underwater Image Enhancement}
The robust representation capabilities of deep learning have opened its potential in the field of UIE. Over time and with technological advancements, data-driven deep learning approaches have emerged as the predominant methodology in UIE \cite{anwar2020diving}. As a pioneering effort, Anwar et al. \cite{anwar2018deep} introduced UWCNN, a CNN-based framework for image enhancement that yielded promising results. Following this, numerous deep learning-based approaches have emerged \cite{anwar2020diving, zhou2023ugif, li2023ruiesr}, employing data-driven strategies to enhance images via end-to-end model training. These methods generally achieve impressive performance by designing sophisticated network architectures, including specialized modules tailored to address specific challenges such as underwater illumination \cite{li2023uialn, zhou2024iacc} and color correction \cite{zhou2023ugif, li2023ruiesr, li2023tctl}, or by integrating UIE tasks with other tasks such as depth estimation \cite{Ye2020Deep, yang2024joint} and object detection \cite{Yeh2022Lightweight, liu2024unitmodule}. More recently, with the development of large models today, researchers have successfully harnessed their power to achieve impressive improvements in UIE tasks. For example, Liu et al. \cite{liu2024underwater} used a diffusion model as the backbone and introduced a CLIP classifier to guide the enhancement process, thereby achieving a more natural visual enhancement effect.

Despite substantial improvements in both visual effects and generalization over traditional methods, these approaches frequently introduce distortions and inappropriate enhancements. This occurs because the robustness and reliability of data-driven methods are highly dependent on access to a large volume of high-quality training data. Moreover, expanding the scale of the model further increases the costs associated with training and learning. However, due to the challenges associated with collecting underwater imagery in various underwater environments and the practical infeasibility of obtaining truly clear, water-free reference images, the availability of high-quality underwater image datasets is significantly constrained. Most of the reference images within these datasets require the initial generation of candidate enhancements through a variety of enhancement techniques, followed by manual selection of the highest-quality images to serve as pseudo references \cite{li2019underwater, Qi2022SGUIE}. Their questionable reliability poses a significant impediment to the advancement of deep learning-based approaches.

\begin{figure*}[t]
  \centering
   \includegraphics[width=1\linewidth]{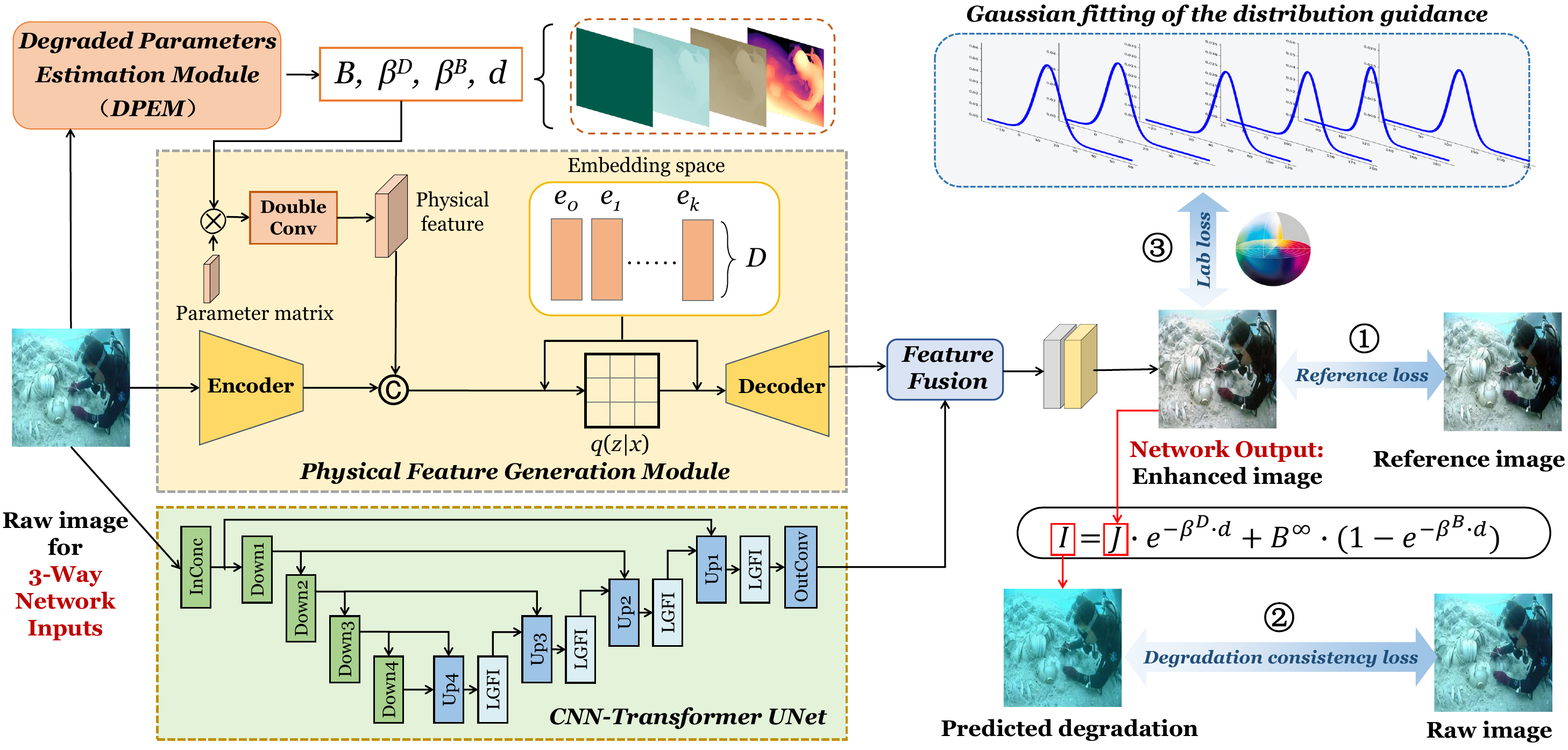}
   \caption{The framework of the proposed DPF-Net. The two branches, CNN-Transformer UNet (CT-UNet) and Physical Feature Generation Module (PFGM), receiving physical parameters input from the pre-trained Degraded Parameters Estimation Module (DPEM) into embedding space. The final output image is obtained after the output features from the two branches undergo the Feature Fusion module. The network loss includes reference loss, degradation consistency loss, and weak reference loss based on the Lab color space.}
   \label{fig:DPF-Net}
\end{figure*}

\subsection{Fusion of Data-driven and Physical Model}
Leveraging the theoretical foundation provided by physical models as constraints and integrating the restoration learning capabilities of data-driven methods is a promising complementary solution. In fact, several studies have explored initial integration strategies by utilizing physical imaging models to simulate underwater degradation in terrestrial images, generating synthetic paired underwater images for data-driven training purposes \cite{chen2020perceptual, wen2023syreanet, wang2023domain}. Furthermore, certain researchers have investigated more advanced fusion methodologies to more effectively leverage their complementary strengths. Wang et al. \cite{wang2021single} proposed an analysis-synthesis framework for UIE tasks, which is trained on a synthesized dataset. They introduced several underwater scene priors, including color tone prior, water type prior, and structural prior, which are integrated into the enhancement process via the Prior Encoder and Guidance Module. However, these priors are artificially constructed based on imaging models and image content, with limited incorporation of physical models in the enhancement process. 

Du et al. \cite{du2024physical} innovatively developed a Deep Degradation Model (DDM) to estimate physical parameters, leveraging degraded images for supervision. However, the DDM is concurrently trained alongside an enhancement model from scratch, which poses challenges because the combined loss function may not adequately supervise the parameters during synchronous training. Consequently, the DDM appears to capture only the correlation between the raw and non-fully-reliable reference images, rather than embodying an authentic physical imaging model, which makes it difficult to achieve optimal results. Therefore, for an enhanced strategy that integrates data-driven methods with physical models, a critical challenge is to effectively incorporate reliable constraints from physical imaging models into the deep learning framework while minimizing potential negative impacts arising from unreliable references within the data-driven approach.

\section{Proposed Method}\label{sec:Pro}

\subsection{Overall Framework}
\begin{figure*}[t]
  \centering
   \includegraphics[width=1\linewidth]{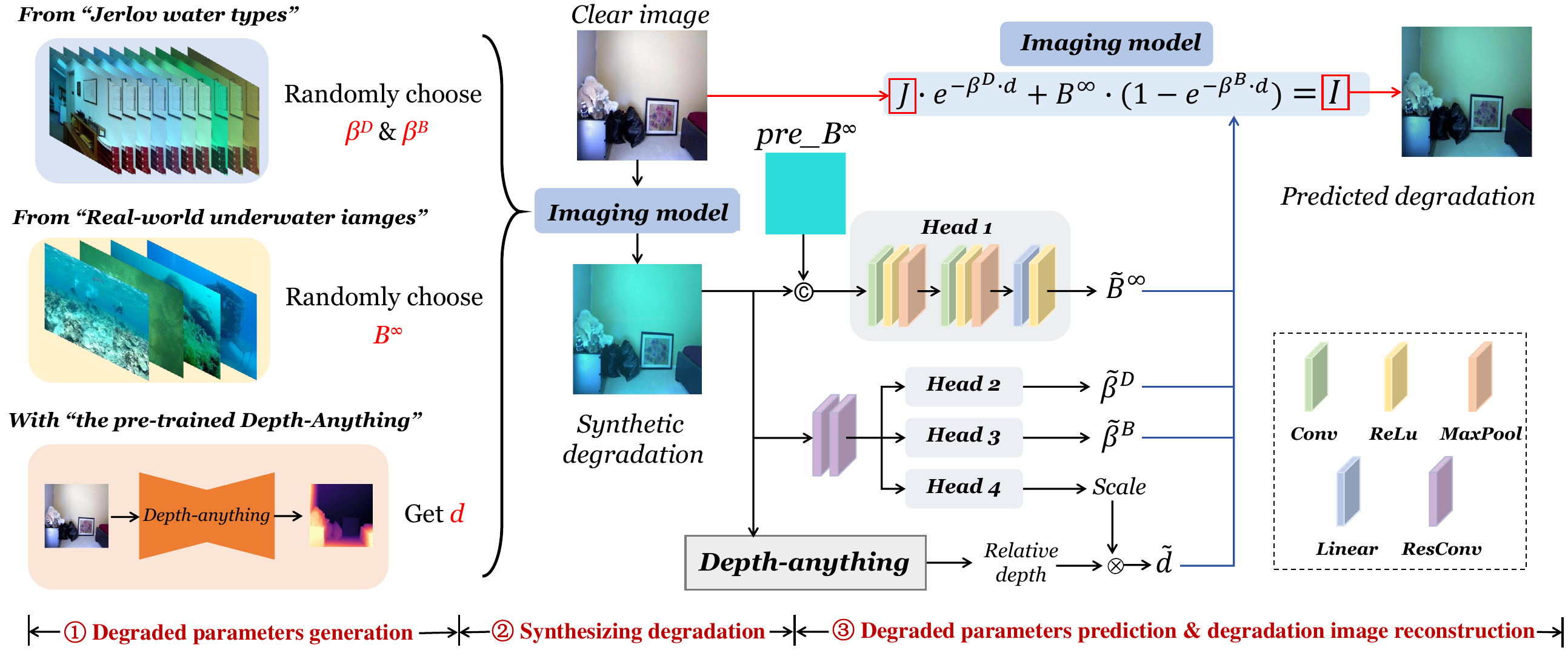}
   \caption{The architecture and training methodology of the proposed DPEM. Underwater images are synthesized from terrestrial RGB-D images by configuring various combinations of degradation parameters within the underwater physical imaging model. Then, DPEM is trained to predict these degradation parameters using the synthesized images as ground truth.}
   \label{fig:DPEM}
\end{figure*}

To address the aforementioned challenges, as shown in Figure \ref{fig:DPF-Net}, we propose an advanced underwater image enhancement network that integrates deep learning techniques with physical models through embedded feature fusion. Its construction and training adopt a two-stage framework. In the first stage, as shown in Figure \ref{fig:DPEM}, we develop a Degraded Parameters Estimation Module (DPEM) and train it on a synthetic dataset using predefined degraded parameters. This ensures that the model's output predictions are consistent with the imaging model. In the second stage, we incorporate the estimated degraded parameters into the feature extraction process of our proposed deep enhancement model, DPF-Net, and utilize the imaging model for degradation reconstruction to provide supervision.

The main image enhancement network, DPF-Net, comprises two branches: an enhanced CNN-Transformer UNet (CT-UNet) that integrates both CNN and Transformer architectures, and a Physical Feature Generation Module (PFGM) built on the Vector Quantized Variational Autoencoder (VQ-VAE) framework. In the first branch, the raw image is processed through a U-Net encoder-decoder architecture to extract CT-UNet features. This process emphasizes feature extraction and purely data-driven refinement within the network. In the second branch, the image is input into the VQ-VAE encoder with a key distinction: it incorporates the pre-trained degraded parameters estimation module to predict the physical parameters of the raw image. These predicted physical parameters are then transformed into feature matrices by matrix multiplication followed by two convolutional layers. The features extracted from both the image and the physical parameters are mapped into an embedding space and further processed by the decoder to generate PFGM features. Finally, the features obtained from the CT-UNet and PFGM branches are integrated via a fusion module, resulting in an enhanced output image.

\subsection{Degraded Parameters Estimation Module}
The primary objective of the Degraded Parameters Estimation Module within our DPF-Net, as illustrated in Fig. \ref{fig:DPF-Net}, is to provide reliable and physically consistent predictions of degradation parameters for underwater images. Fig. \ref{fig:DPEM} illustrates the architecture and training methodology of DPEM, which roughly consists of three main steps: degraded parameters generation, degradation synthesis, and degraded parameters prediction and degradation image reconstruction. DPEM is an independently developed and trained module. We first synthesize underwater images using terrestrial RGB-D images by configuring different combinations of degraded parameters in the underwater physical imaging model, and subsequently train DPEM to predict the degraded parameters using these preset parameters as ground truth.

\subsubsection{Degraded Parameters Generation and Degradation Image Synthesizing}
\begin{figure}[t]
  \centering
   \includegraphics[width=1\linewidth]{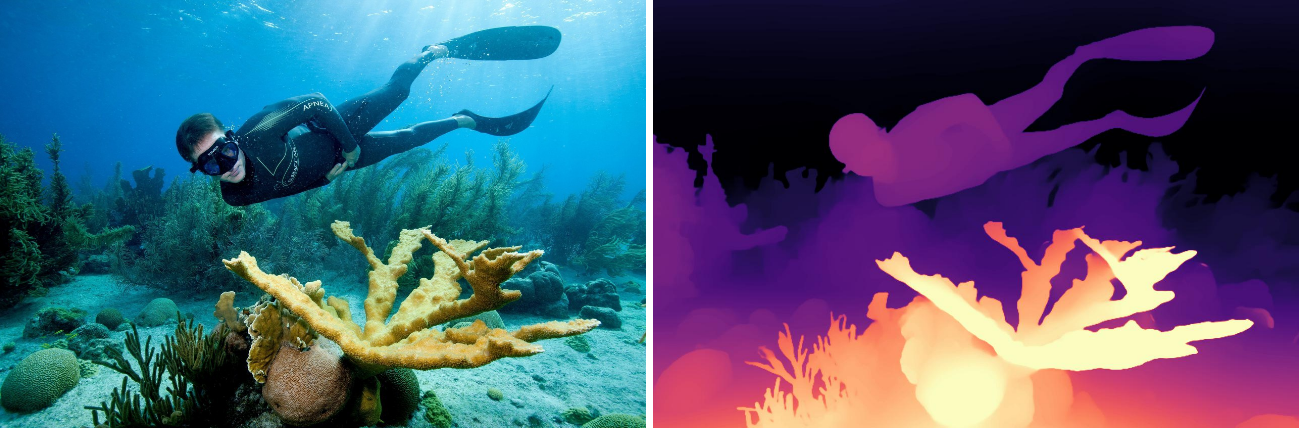}
   \caption{Raw image (left) and corresponding relative depth map (right).}
   \label{fig:raw_depth}
\end{figure}

We utilize NYU-Depth-V2 \cite{silberman2012indoor}, which included $1,449$ indoor RGB-D images, to synthesize the training dataset. This dataset includes absolute depth data, with a maximum depth of approximately $10$ meters, and the relatively limited indoor range is also appropriate for underwater photography. Nonetheless, the absolute depth map derived from the binocular reconstruction of NYU-Depth-V2 lacks accuracy; hence, only the maximum and lowest values of the dataset are used as the absolute depth scale. We use Depth-anything \cite{yang2024depth}, which is a large model for monocular depth estimation, to obtain the relative depth map and subsequently multiply it by the absolute depth scale to derive the absolute depth map. Figure \ref{fig:raw_depth} illustrates an example of a relative depth map where warmer colors indicate proximity in relative depth, while cooler colors signify greater distance.

\begin{figure}[t]
  \centering
   \includegraphics[width=1\linewidth]{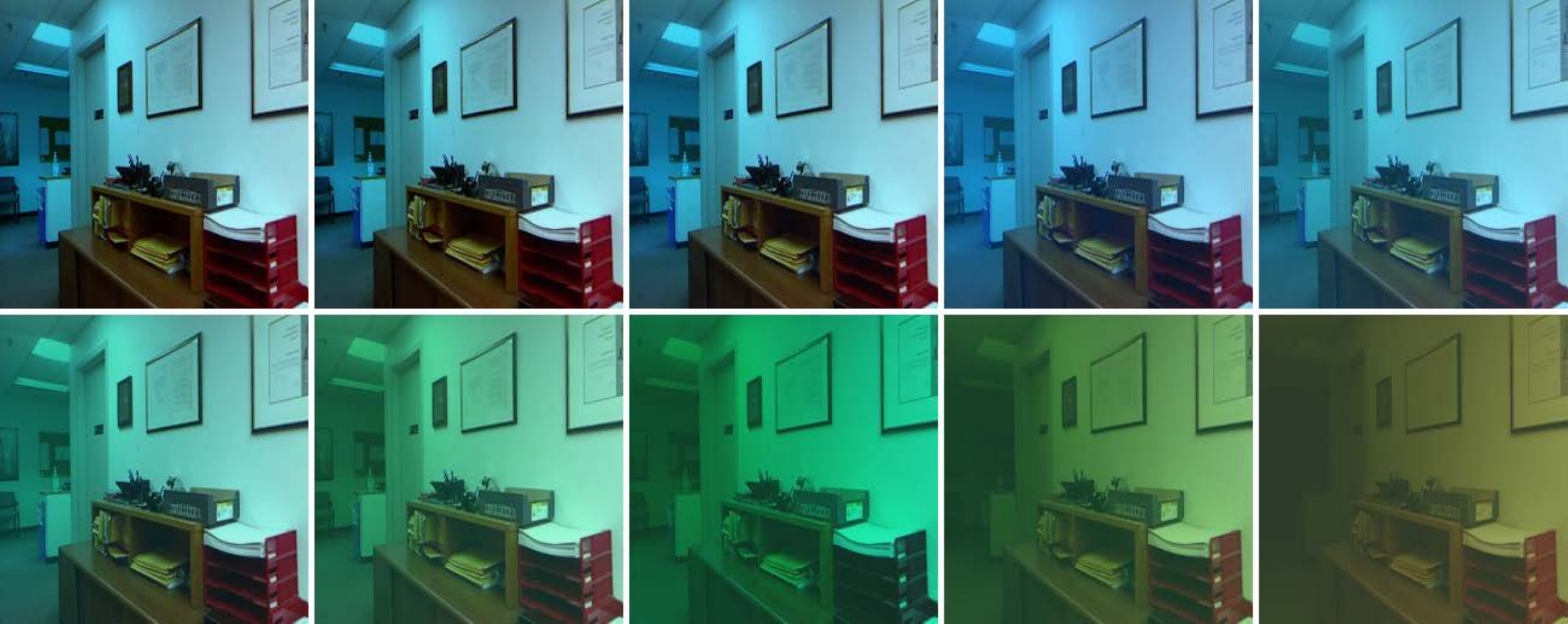}
   \caption{Different synthetic degradation images of one in NYU-Depth-V2 \cite{silberman2012indoor}.}
   \label{fig:ten_Jerlov}
\end{figure}

For other degraded parameters required by Eq. (\ref{eq1}), i.e., $\beta^D$, $\beta^B$ and $B^{\infty}$, we can generate different combinations to synthesize the corresponding degradation images for various water types, ensuring that they conform to the physical model of underwater imaging. As mentioned in Section. \ref{sec:Rel_A}, we selected the attenuation and scattering coefficients (from \cite{solonenko2015inherent}) with wavelengths of 650 nm, 525 nm, and 450 nm to serve as the $\beta^D$ and $\beta^B$ values for the RGB channels, respectively. A pair of these types is randomly designated as $\beta^D$ and $\beta^B$ for one synthetic degradation image. Then, we randomly select an image from the real underwater dataset and employ Eq. (\ref{eq3}) to independently calculate the background light of the Green channel, denoted as $B^{\infty}_g$. According to Zhao et al. \cite{zhao2015deriving}, $B^{\infty}$ is inversely proportional to $\beta^D$ and directly proportional to $\beta^B$. With $\beta^D$, $\beta^B$, and $B^{\infty}_g$ established, the $B^{\infty}$ used for synthesis can be expressed as follows:
\begin{align}  
    B^{\infty} &= \{B^{\infty}_r, B^{\infty}_g, B^{\infty}_b\} \nonumber \\  
               &= \left\{\frac{\beta^D_g \cdot \beta^B_r}{\beta^D_r \cdot \beta^B_g} \cdot B_g^{\infty}, B_g^{\infty}, \frac{\beta^D_g \cdot \beta^B_b}{\beta^D_b \cdot \beta^B_g} \cdot B_g^{\infty}\right\},  
    \label{eq4}  
\end{align}
which ultimately forms the diverse combinations of degraded parameters. Using these combinations of degraded parameters and their interrelations, along with the absolute depth map mentioned above, we can generate synthetic underwater degradation images that conform to both the imaging model and human perception as described in Eq. (\ref{eq1}). Figure \ref{fig:ten_Jerlov} illustrates the synthetic degradation images for ten different water types.

\subsubsection{Degraded Parameters Prediction and Degradation Image Reconstruction}

\begin{figure}[t]
  \centering
   \includegraphics[width=1\linewidth]{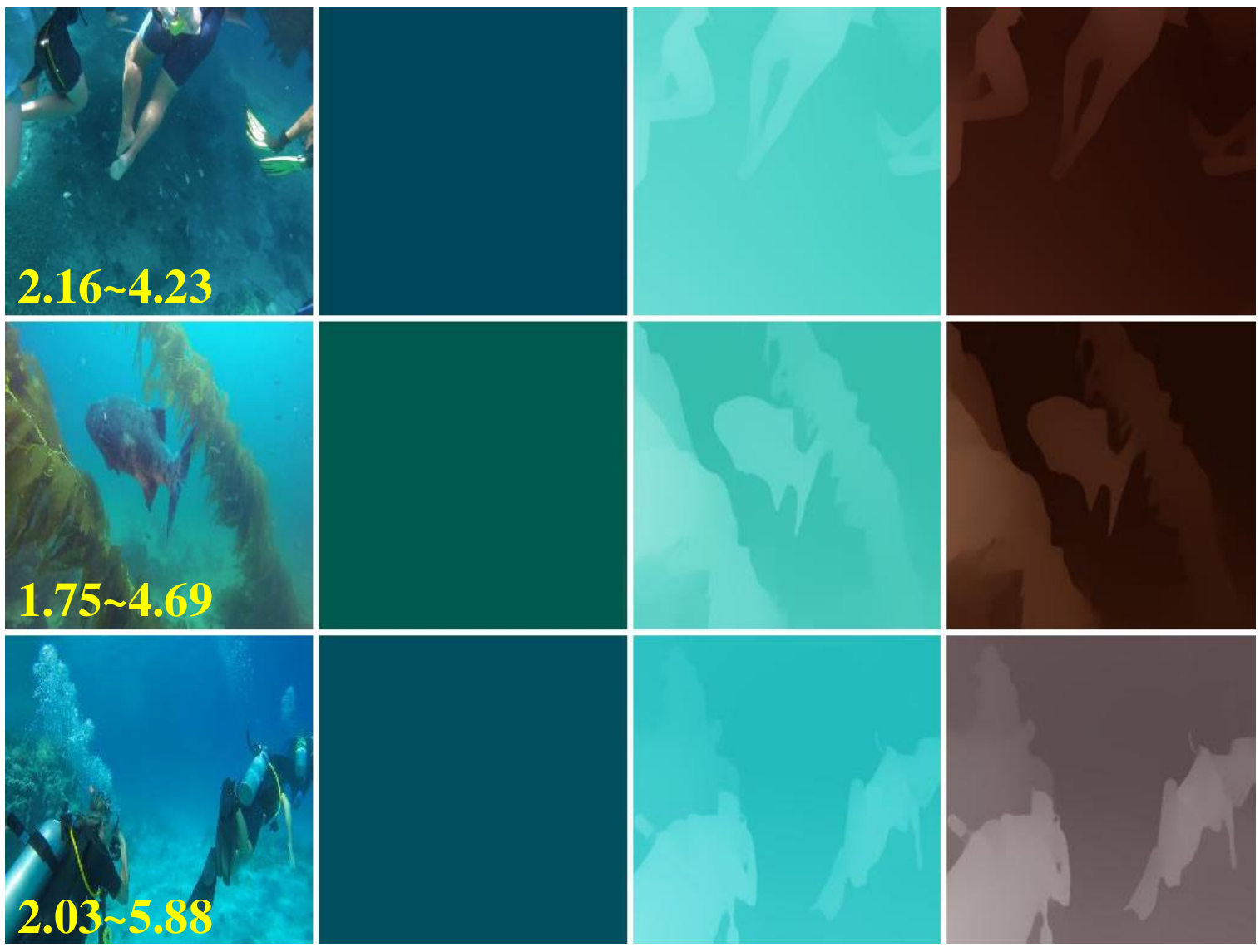}
   \caption{Visualization of degraded parameters predicted by DPEM, with each row showing (from left to right): the raw image, $B^{\infty}$, $255 \cdot e^{-\beta^D \cdot d}$, and $255 \cdot e^{-\beta^B \cdot d}$. The lower left figures indicate the scale of the absolute depth.}
   \label{fig:DPEM_output}
\end{figure}

Having outlined the physical parameters required for synthesis, we now focus on the network architecture and the training strategy of the DPEM. For $B^{\infty}$, we first use Eq. (\ref{eq2}) and Eq. (\ref{eq3}) to calculate the preliminarily estimated background light $pre\_B^{\infty}$ of the synthetic image, and concatenate it with the image. Then the concatenated data with 4 channels is fed into a prediction head composed of convolutional layers to get the predicted value of $B^{\infty}$. For the other parameters, the synthetic image is directly entered into a common encoder composed of shared residual convolutional layers for initial feature extraction. Then, different prediction heads yield $\beta^D$, $\beta^B$, and $Scale$, respectively. $Scale$ represents the minimum and maximum values of the absolute depth scale of the image, which must be multiplied by the relative depth map derived by Depth-anything \cite{yang2024depth} to obtain the absolute depth map $d$. The supervision of the outputs of each prediction head comes from the degraded parameters used in the synthetic image generation process, which serve as the labels for model training. Furthermore, the same degradation process is applied to the clear image according to Eq. (\ref{eq1}) using the predicted physical parameters to obtain a predicted degraded image, which is used to calculate the degradation consistency loss.

\subsubsection{The Training and Performance Overview of DPEM}
The loss of DPEM training consists of two components: parameter loss and degradation consistency loss. The parameter loss evaluates the L1 distance between the predicted parameters and the ground truth, ensuring the precision of the output parameters generated by each prediction head. The degradation consistency loss comprises both the VGG16-based perceptual loss and the Structural Similarity Index (SSIM) loss, which are calculated between the predicted degradation image and the synthetic degradation image. Specifically, the former quantifies the perceptual similarity between the two images by utilizing features extracted from the pre-trained VGG16 network. SSIM loss assesses the similarity between predicted and synthetically degraded images based on their luminance, contrast, and structural information. Incorporating image-level losses can enhance the convergence of the overall loss function during network training.

The absence of truth values for physical parameters in real underwater images precludes the evaluation of DPEM's output. However, these parameters can be visualized using Eq. (\ref{eq1}). Figure \ref{fig:DPEM_output} illustrates the visible results of several test examples, with each row progressing from left to right to represent the raw image, $B^{\infty}$, $255 \cdot e^{-\beta^D \cdot d}$, and $255 \cdot e^{-\beta^B \cdot d}$. The figures in the lower left corner of the raw image denote the predicted absolute minimum and maximum depth scales.

\begin{figure}[t]
  \centering
   \includegraphics[width=1\linewidth]{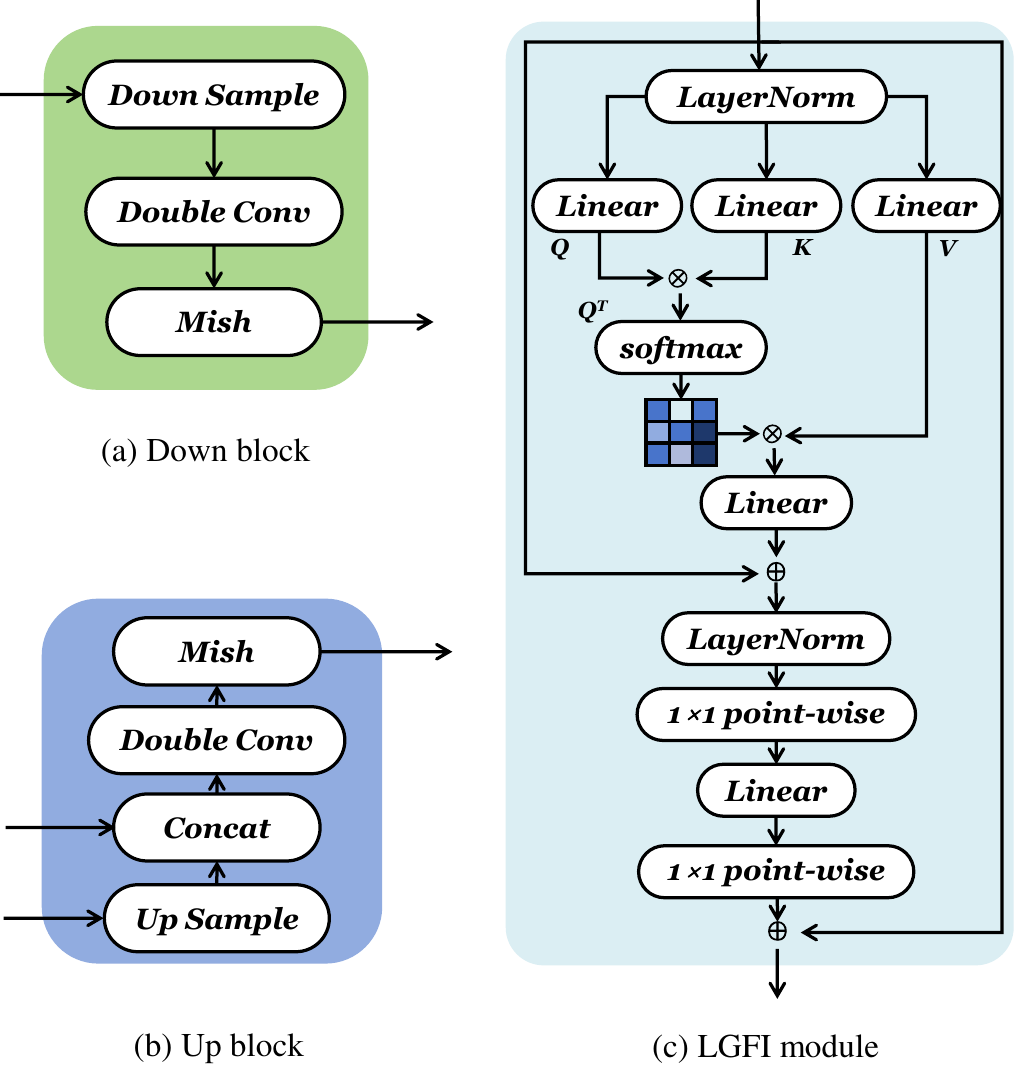}
   \caption{The structure of (a) Down block, (b) Up block, and (c) the LGFI module of the CT-UNet.}
   \label{fig:CT-UNet}
\end{figure}

\subsection{Data Driven and Physical Parameters Fusion Network}
In addition to the DPEM module introduced above, the proposed UIE framework DPF-Net comprises two primary branches: an improved UNet that employs both CNN and Transformer structures and a Physical Feature Generation Module (PFGM). The architecture of CT-UNet is depicted in the lower branch of Figure \ref{fig:DPF-Net}, with detailed illustrations of the Up block, Down block, and Local-Global Features Interaction (LGFI) module provided in Figure \ref{fig:CT-UNet}. This architecture builds upon the conventional UNet framework, utilizing a four-layer encoder-decoder structure that facilitates feature extraction from input images. The input image dimensions are $H \times W \times 3$, from which shallow features of size $H \times W \times C_1$ are extracted via convolutional operations at the input layer. These features are then progressively downsampled through subsequent Down blocks, resulting in feature maps of $\frac{H}{2^n} \times \frac{W}{2^n} \times C_n$, where $n$ represents the downsampling factor at the $n$-th layer.

Subsequently, the Up blocks perform upsampling on the output of the preceding layer and concatenate this upsampled result with the corresponding output from the associated Down block. The concatenated result is then processed through convolutional decoding. While CNNs excel in capturing local information, extracting global features typically requires larger convolutional kernels, which can lead to a more complex model and diminished processing efficiency. To address this limitation, we introduce the LGFI module \cite{zhang2023lite} into our framework, which is based on the Transformer architecture. This module enhances feature extraction capabilities while maintaining the model's lightweight nature and computational speed. In our CT-UNet, $3 \times 3$ convolutions are utilized to extract local features from the input image, followed by the application of the LGFI module in the decoder stage to dynamically integrate local and global information. The final CT-UNet features are generated from the convolutional layer in the last stage.

The upper middle section of Figure \ref{fig:DPF-Net} illustrates the architecture of the PFGM for physical feature generation based on VQ-VAE. The input image is first processed by the VQ-VAE encoder to extract features, denoted $z_e(x)$, which are subsequently mapped in the embedding space. Simultaneously, the raw image is fed into our pre-trained DPEM to obtain predicted physical degraded parameters. These parameters are then element-wise multiplied with the feature matrix. After passing through two convolutional layers, the resulting features, designated as $z_e(p)$, are concatenated with $z_e(x)$ and introduced into the embedding space. The degraded parameters of various water types exhibit fundamentally distinct characteristics, and VQ-VAE can effectively learn these discrete vectors. By leveraging discrete representations in the embedding space, the network is capable of capturing degradation information that is specific to different water types. Finally, the concatenated features yield $z_q(x)$, which undergoes further processing by the decoder to generate the PFGM output features.

\begin{figure}[t]
  \centering
   \includegraphics[width=1\linewidth]{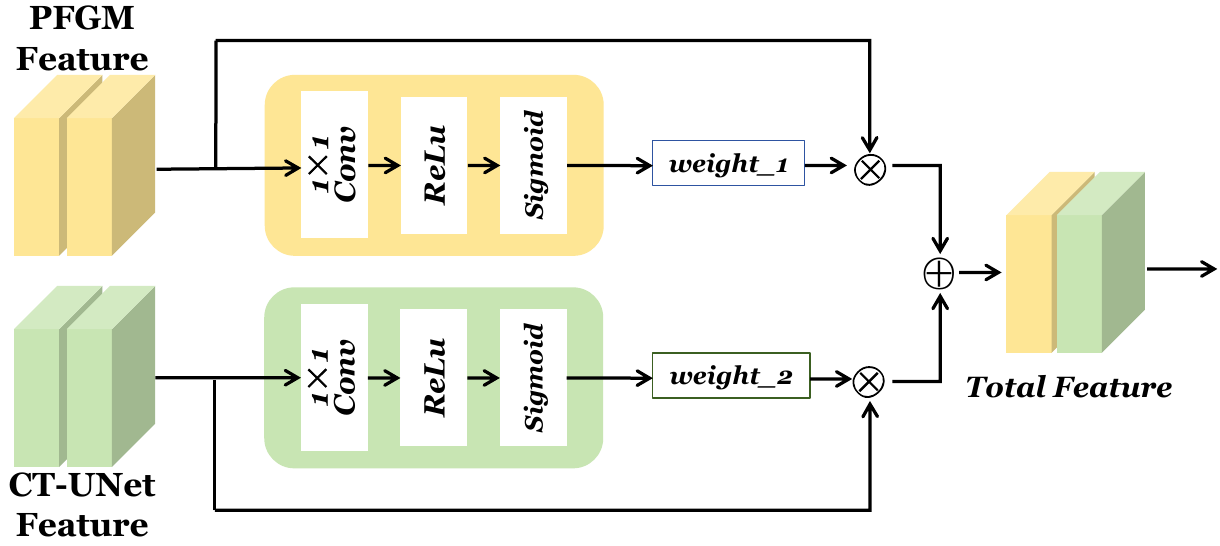}
   \caption{The framework of the feature fusion module.}
   \label{fig:feature_fusion}
\end{figure}

To extract degraded image features that are integrated with physical imaging priors, the feature sets derived from the aforementioned two branches are fed into the feature fusion module depicted in Figure \ref{fig:feature_fusion} for comprehensive integration. Two fusion weights are adaptively derived through convolution operations, following which the features are multiplied by their respective weights for aggregation. The significance of the feature fusion module lies in its integration of learnable weights, which enable the network to determine the relative importance of features of the two branches during training. This process optimizes the utilization of relevant information. Ultimately, the fused features are input into the final convolutional layer to produce the output of DPF-Net.

Within the DPF-Net architecture, CT-UNet plays a crucial role in image feature extraction by utilizing a CNN-Transformer-based network, which significantly enhances the model's learning capabilities. Concurrently, PFGM integrates both image data and physical parameters into the embedding space, which is vital for improving the model's robustness and the effectiveness of physical parameter enhancement. The synergy between these two components ensures optimal performance and adaptability of the DPF-Net.

\subsection{Loss Function}
For deep networks designed for the UIE task, establishing an appropriate loss function is crucial to effective training. For DPF-Net, the challenge lies in designing the loss function to harmonize physical imaging prior constraints with data-driven constraints. To achieve this, we have developed a novel composite loss function as follows: 
\begin{equation}
    L_{DPF-Net} = \alpha_1 \cdot L_{ref} + \alpha_2 \cdot L_{deg} + \alpha_3 \cdot L_{Lab},
    \label{eq5}
\end{equation}  
where the three components are: regionally weighted reference loss $L_{ref}$, degradation consistency loss $L_{deg}$, and weakly supervised reference loss $L_{Lab}$. The weights assigned to each term ($\alpha_n$) are specifically set to 0.6, 0.2, and 0.2 for $\alpha_1$, $\alpha_2$, and $\alpha_3$, respectively.

\subsubsection{Regionally Weighted Reference Loss}
Paired underwater enhancement samples from publicly available datasets, such as UIEB, serve as valuable references for training data-driven models. However, these datasets also contain a significant number of unreliable references or local defects. Specifically, regions farther from the camera tend to exhibit more severe degradation, and the restoration provided by the reference images in these areas is less reliable. It is reasonable to relax the pixel-level constraints imposed by distant references on enhanced images, taking into account the depth relationships within the image scene. Therefore, we introduce a modified reference-based loss, namely the regionally weighted reference loss:
\begin{equation}
    L_{ref} = d_{rel} \cdot L1\left(I_{enc}, I_{ref}\right) + L_{SSIM}\left(I_{enc}, I_{ref}\right),
    \label{eq6}
\end{equation}
where $I_{enc}$ and $I_{ref}$ denote model output enhancement and the reference, respectively. Here, $L1$ denotes the L1 loss between the enhanced output and the reference image, calculated as the mean absolute difference of the pixel values. $d_{rel}$ represents the normalized relative depth map, where closer regions are assigned higher values, thus increasing their contribution to the loss. Multiplication by $d_{rel}$ ensures that the influence of distant regions is reduced. Additionally, $L_{SSIM}$ denotes the SSIM loss between the reference image and the output, quantifying the structural similarity and assessing the disparity between the two.

\subsubsection{Degradation Consistency Loss}
Degradation consistency loss $L_{deg}$ can be described in detail by the following equation:
\begin{equation}
    L_{deg} = L_{SSIM}\left(I_{raw}, I_{deg}\right) + L_{VGG}\left(I_{raw}, I_{deg}\right),
    \label{eq7}
\end{equation} 
which incorporates the SSIM loss $L_{SSIM}$ and the perceptual loss $L_{VGG}$ utilizing the pre-trained VGG16 network, between the raw image $I_{raw}$ and the synthetic predicted degradation image $I_{deg}$ obtained through physical parameters and the output enhanced image. The introduction of this loss function enables the effective integration of physical knowledge-guided and data-driven enhancement strategies, thereby facilitating their mutual support and constraint. The rationale for not selecting the pixel-to-pixel L1 loss for measuring the discrepancy between the raw image and the predicted degradation image is to prevent imposing overly stringent constraints on the resultant enhanced image generated by the physical model. This decision is driven by the suboptimal outcomes commonly observed in the physical enhancement framework. Introducing pixel-level losses, such as $L1$, may adversely affect the model's enhancement efficacy, perhaps resulting in color block distortion in the enhanced images during training.

\subsubsection{Weakly Supervised Reference Loss}
\begin{figure}[t]
  \centering
   \includegraphics[width=0.95\linewidth]{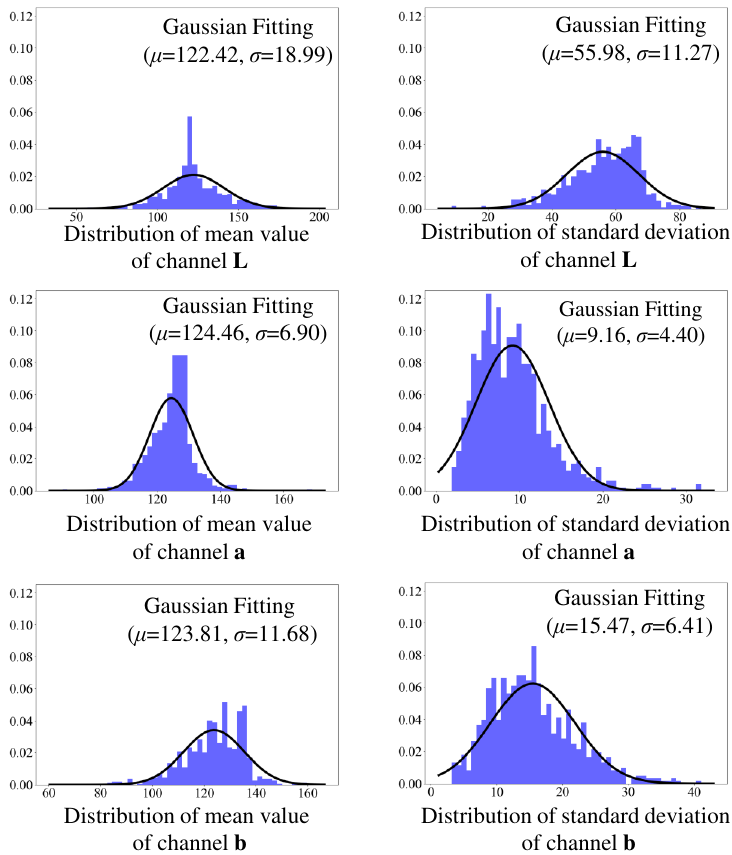}
   \caption{Distributions of the mean values and standard deviations for the three channels of reference images in the UIEB training set, represented in the Lab color space. Gaussian curves are utilized to fit these distributions. The left column represents the distributions of mean values of the three channels, while the right column shows those of the standard deviations. The X-axis denotes the specific values of the channel, which are divided into 100 bins, and the Y-axis represents the proportion of samples within each bin relative to the entire dataset.}
   \label{fig:Lab}
\end{figure}

Motivated by color transfer techniques that operate within the Lab color space for image color adaptation \cite{reinhard2001color}, our proposed weakly supervised reference loss $L_{Lab}$ aims to enhance the appearance of the image color to better align with the color distribution of the reference dataset. This approach avoids the limitations imposed by relying on a single paired reference image, which may hinder further network learning due to potential quality issues in the reference. In Figure \ref{fig:Lab}, we present the distributions of the mean values and standard deviations for the three channels of reference images in the UIEB training set, represented in the Lab color space. Gaussian curves are utilized to fit these distributions. We can notice that the statistics of color appearance are notably concentrated, which inspires us that the above-mentioned statistics of the enhanced image we output should not significantly deviate from the above distribution. Constraining it as a loss function can facilitate the alignment of the output with the distribution of the overall reference dataset, while enhancing color vividness and preventing severe distortion during the enhancement process.

Specifically, in the computation of $L_{Lab}$, the enhanced image is first transformed into the Lab color space. Subsequently, the mean $\mu_c$ and the standard deviation $\sigma_c$ are calculated for each of the three channels. It is expected that the color statistics will converge to the pre-determined Gaussian distribution. Therefore, the Gaussian Probability Density Function (GPDF) is utilized for loss computation:
\begin{equation}
    L(x) = \frac{(x-\mu_n)^2}{2\sigma^2_n},
    \label{eq8}
\end{equation}
where $x$ denotes either the mean value $\mu_c$ or the standard deviation $\sigma_c$ of channel $c$ in the enhanced image within the Lab color space, whereas $\mu_n$ and $\sigma_n$ represent the mean and standard deviation corresponding to the six Gaussian curves, respectively. The aforementioned GPDF is computed six times, aggregating the losses with equal weights to derive $L_{Lab}$. $L_{Lab}$ represents a weak supervision loss that depends on the entire reference set. In practice, the requirements for the reference set are relatively lenient; even clear terrestrial images can be utilized as a dataset to fit six Gaussian curves. As long as an appropriate color distribution is established in the Lab space, the model output can be effectively guided.

\section{Experiment}\label{sec:Expe}
\subsection{Implementation Details and Experimental Setting}\label{sec:details}
Initially, we trained the DPEM model on the NYU-Depth-V2 dataset for $50$ epochs. Subsequently, we trained our DPF-Net and sequentially fine-tuned the DPEM model. For DPF-Net, the initial learning rate was set to $1e^{-3}$ and then reduced by a factor of $0.9$ every 10 epochs. For DPEM, the learning rate was initialized at $5e^{-6}$, which is significantly smaller than the initial learning rate of DPF-Net, in order to fine-tune DPEM rather than train it from scratch. During the fine-tuning phase, only the two DPEM prediction heads, responsible for the prediction of the parameters $\beta^D$ and $\beta^B$, were updated. Both DPF-Net and the relevant parameters of DPEM were optimized using the Adam optimizer. All model training, testing, and associated experiments were conducted on a server equipped with an NVIDIA RTX 3090 GPU, an Intel Xeon Silver 4210R processor operating at 2.40GHz, and running the Ubuntu 18.04 operating system.

\subsection{Datasets and Evaluation Metrics}
\subsubsection{Datasets} We employed the widely recognized UIEB dataset \cite{li2019underwater} for training, which consists of $890$ natural underwater images paired with manually selected reference images. Specifically, $800$ images were randomly chosen for training purposes, while the remaining $90$ images were designated as the test set, referred to as \textbf{UIEB-T}. In addition to being used for the training of DPF-Net, images from the UIEB training set were also used to randomly determine the values of the $B^{\infty}$ parameters for degradation synthesis during the training of DPEM.

To perform a more comprehensive evaluation and accurately assess the performance of the model under challenging underwater conditions, we evaluated the model using the \textbf{UIEB-Challenging} and test set from the \textbf{SUIM-E} \cite{Qi2022SGUIE} datasets. The UIEB-Challenging dataset comprises $60$ underwater images characterized by severely degraded imaging quality and blurriness. The SUIM-E dataset includes pre-divided $1,525$ pairs of training images and $110$ pairs of test images; only the test set was utilized to evaluate comparison methods. Additionally, to evaluate the generalization capabilities of different approaches on underwater images with varying degrees of degradation, we conducted further experiments on the \textbf{RUIE} \cite{liu2020real} dataset, which is widely adopted as a no-reference test set. Specifically, we employed the Underwater Image Quality Set (UIQS) from this dataset, categorized into five subsets labeled A through E based on the extent of degradation from low to high. To avoid redundancy from successive scenes, we extracted $20$ images at uniform intervals from each subset. Images sampled from subsets A, B, and C were consolidated to create a collection exhibiting low-level degradation, referred to as \textbf{RUIE-ABC60}. Images sampled from subsets D and E were combined to form a collection characterized by extreme degradation, referred to as \textbf{RUIE-DE40}.

\subsubsection{Evaluation Metrics} 
We employed four evaluation metrics for both reference-based and non-reference-based assessments: the Peak Signal-to-Noise Ratio (PSNR), the Structural Similarity Index (SSIM) \cite{wang2004image}, the Underwater Image Quality Metric (UIQM) \cite{panetta2015human}, and the Underwater Color Image Quality Evaluation (UCIQE) \cite{yang2015underwater}. PSNR quantifies image quality by measuring pixel-wise differences between the output and reference images, with higher values indicating greater similarity. SSIM evaluates visual quality based on luminance, contrast, and structure, with higher scores reflecting a closer alignment between the enhanced and reference images. UIQM and UCIQE are non-reference-based metrics. UIQM assesses three key attributes: underwater contrast, sharpness, and colorfulness. UCIQE is designed to detect perceptible changes in image quality as perceived by the human visual system, generating a single quality score through statistical analysis of color distribution and contrast.

\begin{figure*}[t]
  \centering
   \includegraphics[width=0.95\linewidth]{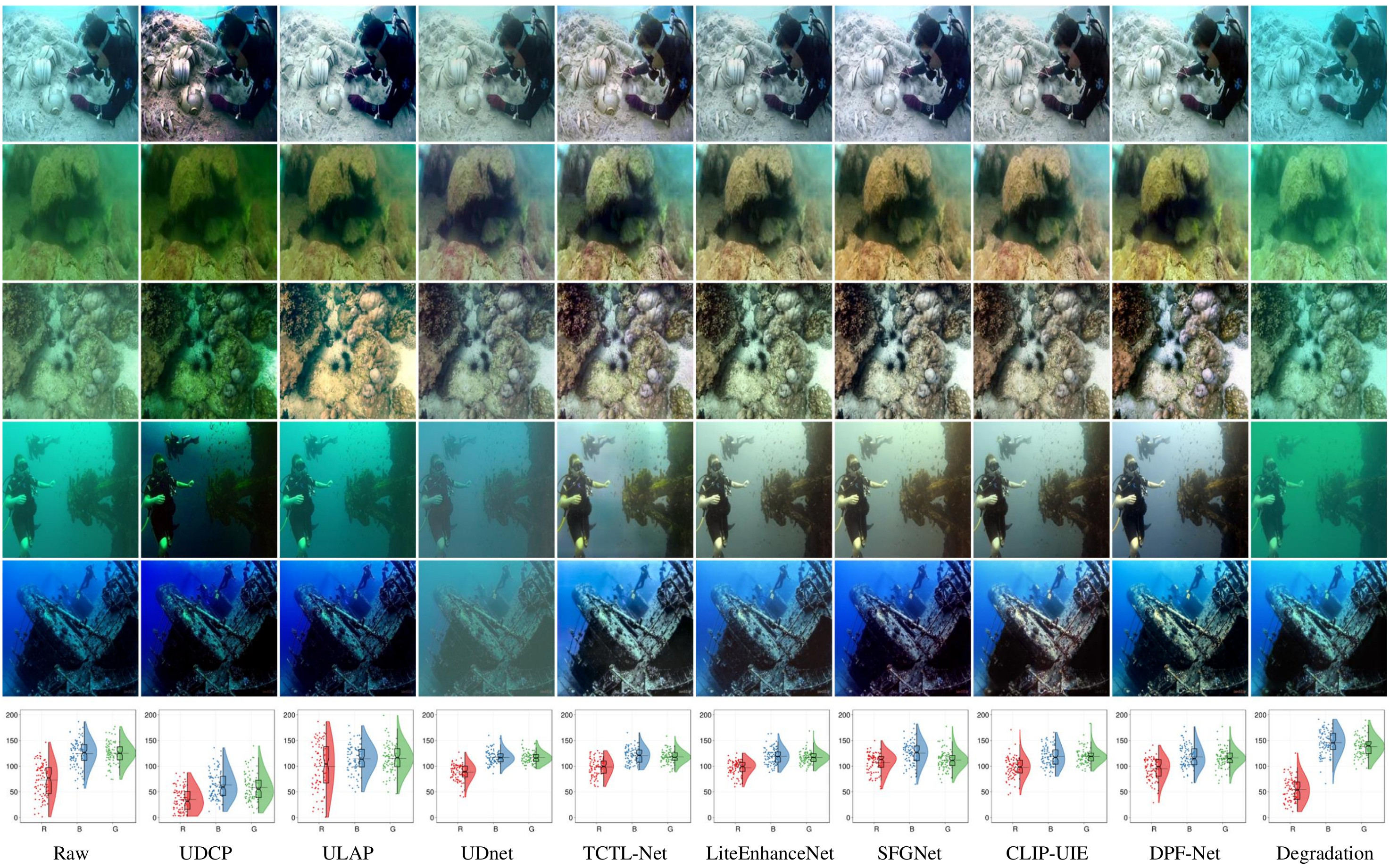}
   \caption{Visual comparisons on images from UIEB-T (the first three rows) and SUIM-E (fourth and fifth rows). From left to right, the figure presents raw underwater images, followed by the results of UDCP \cite{drews2016underwater}, ULAP\cite{song2018rapid}, UDnet\cite{saleh2022adaptive}, TCTL-Net\cite{li2023tctl}, LiteEnhanceNet \cite{zhang2024liteenhancenet}, SFGNet\cite{zhao2024toward}, CLIP-UIE\cite{liu2024underwater}, the proposed DPF-Net, reference images, and the predicted degradation images, respectively. The bottom row illustrates the statistical distribution of the mean values across the three RGB channels for enhanced images within the entire UIEB-T dataset.}
   \label{fig:UIEB&SUIM_result}
\end{figure*}

\begin{table*}[t]
  \centering\small
  \caption{Evaluation results on the UIEB-T and SUIM-E datasets. The best and second-best scores are in red and blue fonts, respectively.}
  \fontsize{9pt}{12pt}\selectfont
  \label{tab:UIEB&SUIM}
   \begin{tabular}{C{0.2\linewidth} | C{0.07\linewidth}  C{0.07\linewidth} C{0.07\linewidth}  C{0.07\linewidth} | C{0.07\linewidth} C{0.07\linewidth} C{0.07\linewidth} C{0.07\linewidth}}
  \Xhline{1pt}
    Datasets &  \multicolumn{4}{c|}{UIEB-T} & \multicolumn{4}{c}{SUIM-E}\\
  \hline
  Methods/Metrics  & PSNR$\uparrow$  & SSIM$\uparrow$ & UIQM$\uparrow$  & UCIQE$\uparrow$ & PSNR$\uparrow$  & SSIM$\uparrow$ & UIQM$\uparrow$ & UCIQE$\uparrow$ \\ 
  \hline
  UDCP\cite{drews2016underwater}   & 11.597 & 0.481 & 1.781 & 0.580 & 11.634 & 0.426 & 1.612 & 0.596 \\
  ULAP\cite{song2018rapid} & 17.945 & 0.697 & 2.378 & \textcolor{red}{0.616} & 16.562 & 0.637 & 2.106 & \textcolor{red}{0.621} \\
  UDnet\cite{saleh2022adaptive} & 21.594 & 0.755 & \textcolor{red}{3.045} & 0.578 & 15.975 & 0.450 & 2.433 & 0.543 \\
  TCTL-Net\cite{li2023tctl} & 22.036 & 0.776 & 3.000 & \textcolor{blue}{0.611} & 20.724 & 0.678 & \textcolor{red}{2.667} & \textcolor{blue}{0.615} \\
  LiteEnhanceNet\cite{zhang2024liteenhancenet} & \textcolor{blue}{23.971} & 0.803 & 2.945 & 0.607 & \textcolor{blue}{21.166} & 0.673 & 2.606 & 0.605 \\
  SFGNet\cite{zhao2024toward} & 23.051 & \textcolor{blue}{0.804} & 2.911 & 0.610 & 21.093 & \textcolor{red}{0.743} & 2.586 & 0.609 \\
  Ours  & \textcolor{red}{25.358} & \textcolor{red}{0.810} & \textcolor{blue}{3.024} & 0.605 & \textcolor{red}{22.775} & \textcolor{blue}{0.722} & \textcolor{blue}{2.613} & 0.608 \\
  \Xhline{1pt}
  \end{tabular}
\end{table*}

\subsection{Compared Methods}
We conduct a comprehensive comparison of our proposed method, DPF-Net, against seven representative approaches. These include two physics-based methods (UDCP \cite{drews2016underwater} and ULAP \cite{song2018rapid}), as well as five data-driven deep learning techniques (UDnet \cite{saleh2022adaptive}, TCTL-Net \cite{li2023tctl}, LiteEnhanceNet \cite{zhang2024liteenhancenet}, SFGNet \cite{zhao2024toward}, and CLIP-UIE \cite{liu2024underwater}). Notably, the latter three methods represent the current state-of-the-art in this field.

UDnet \cite{saleh2022adaptive} is an unsupervised learning approach that incorporates an adaptive uncertainty distribution for enhancement via a conditional Variational Autoencoder (cVAE). TCTL-Net \cite{li2023tctl} proposes a template-free color transfer learning framework to predict transfer parameters, supplemented by attention-driven modules that facilitate the learning of differentiated transfer parameters for enhanced flexibility and robustness. LiteEnhanceNet \cite{zhang2024liteenhancenet} introduces a streamlined network architecture leveraging depthwise separable convolutions as its core component to achieve a balance between performance and computational efficiency. SFGNet \cite{zhao2024toward} employs a two-stage framework based on spatial-frequency interaction and gradient maps. CLIP-UIE \cite{liu2024underwater} utilizes the cutting-edge diffusion models as its foundation and innovatively integrates CLIP as a classifier, focusing on enhancements that emulate natural environmental imaging.

We utilized the identical UIEB dataset split for training, with the exception of the two physical approaches (which require no training) and the diffusion model CLIP-UIE (employing the pre-trained model supplied by the authors). Both the training and the testing stages use exclusively the $256 \times 256$ image size. Furthermore, all comparative methods utilized the source code released by their individual authors to produce results and consistently maintained the same experimental framework throughout all evaluation procedures.

\subsection{Performance Comparison and Analysis}
The quantitative evaluations of compared methods on the UIEB-T and SUIM-E test sets, both of which include paired reference images, are presented in Table \ref{tab:UIEB-SUIM}. Given that the training of CLIP-UIE \cite{liu2024underwater} utilized the UIEB dataset partitioned by the authors as well as the training set from SUIM-E, a potential overlap in the data could introduce bias and render comparisons with other methods unfair. Consequently, we opted not to evaluate the performance of CLIP-UIE on these two test sets. In the evaluation of four reference-based metrics (that is, PSNR and SSIM) across the two aforementioned datasets, our approach achieved three best scores and one second best score.  Given that the reference images for the two datasets were selected through a volunteer-based consensus process, this suggests that the enhancement results obtained using the proposed method are more consistent with human-perceived high-quality visual imaging. Furthermore, based on the results of non-reference-based metrics, DPF-Net demonstrated commendable performance, ranking the second position on the UIQM metric.

To more intuitively illustrate the image enhancement effects of each method, Figure \ref{fig:UIEB&SUIM_result} provides representative comparisons of the UIEB-T and SUIM-E test sets. The two traditional physical methods exhibit noticeable color distortion and loss of detail, leading to suboptimal enhancement results. The unsupervised approach UDnet also demonstrates inadequate enhancement, particularly in terms of low contrast. Deep learning-based competitors exhibit marginally lower visual quality than DPF-Net. Our model has achieved superior visual outcomes, particularly in scenarios involving common types of water degradation and consistent background illumination. Constraining physical parameters notably enhances the robustness of the results, producing output images with natural coloration. The rightmost column in the figures showcases the predicted degradation images generated from the physical parameters forecasted by DPEM, as well as the enhanced images produced by DPF-Net through the physical imaging model formulated in Eq. (\ref{eq1}). These predicted degradation images closely match the raw images in terms of color and detail, thereby validating the reliability of the DPEM physical parameters that we introduced.

\begin{figure*}[t]
  \centering
   \includegraphics[width=0.95\linewidth]{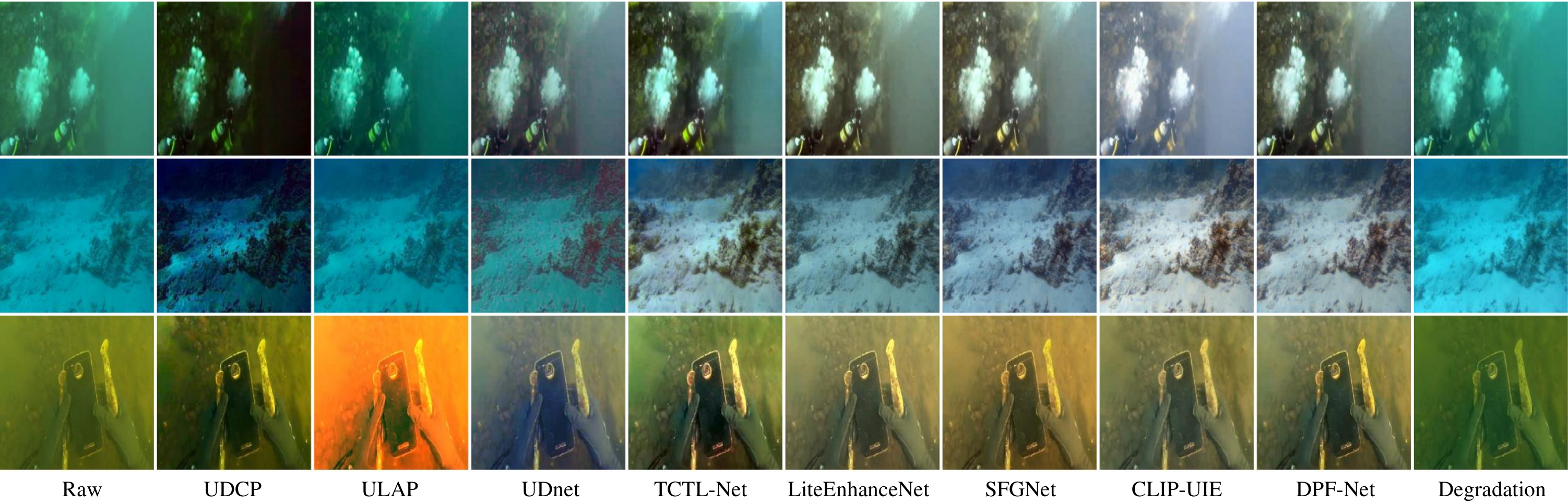}
   \caption{Visual comparisons on images from the UIEB-Challenging dataset. From left to right, raw underwater images and the results of UDCP\cite{drews2016underwater}, ULAP\cite{song2018rapid}, UDnet\cite{saleh2022adaptive}, TCTL-Net\cite{li2023tctl}, LiteEnhanceNet\cite{zhang2024liteenhancenet}, SFGNet\cite{zhao2024toward}, CLIP-UIE\cite{liu2024underwater}, the proposed DPF-Net and the predicted degradation images are presented, respectively.}
   \label{fig:C60_result}
\end{figure*}

\begin{table*}[t]
  \centering\small
  \caption{Evaluation results on the UIEB-Challenging, RUIE-ABC60, and RUIE-DE40 test sets. The best and second-best scores are in red and blue, respectively.}
  \fontsize{9pt}{12pt}\selectfont
  \label{tab:C60&RUIE}
   \begin{tabular}{C{0.2\linewidth} | C{0.07\linewidth}  C{0.07\linewidth} | C{0.07\linewidth}  C{0.07\linewidth} | C{0.07\linewidth}  C{0.07\linewidth}}
  \Xhline{1pt} 
    Datasets & \multicolumn{2}{c|}{UIEB-Challenging} & \multicolumn{2}{c|}{RUIE-ABC60} & \multicolumn{2}{c}{RUIE-DE40} \\
  \hline
  Methods/Metrics    & UIQM$\uparrow$  & UCIQE$\uparrow$ & UIQM$\uparrow$  & UCIQE$\uparrow$ & UIQM$\uparrow$  & UCIQE$\uparrow$ \\
  \hline
  UDCP \cite{drews2016underwater} & 1.460 & 0.646 & 2.288 & 0.537 & 2.026 & 0.524 \\
  ULAP \cite{song2018rapid} & 1.713 & 0.566 & 2.566 & 0.539 & 2.336 & 0.520 \\
  UDnet \cite{saleh2022adaptive} & 2.591 & 0.547 & 2.818 & 0.512 & 2.412 & 0.529 \\
  TCTL-Net \cite{li2023tctl} & \textcolor{red}{2.748} & \textcolor{blue}{0.585} & 3.054 & 0.548 & \textcolor{red}{2.876} & \textcolor{red}{0.571} \\
  LiteEnhanceNet \cite{zhang2024liteenhancenet} & 2.599 & 0.552 & \textcolor{blue}{3.067} & 0.551 & 2.798 & 0.545 \\
  SFGNet \cite{zhao2024toward} & 2.495 & 0.564 & 3.021 & 0.554 & 2.789 & 0.547 \\
  CLIP-UIE \cite{liu2024underwater} & 2.574 & \textcolor{red}{0.588} & 2.991 & \textcolor{red}{0.574} & 2.778 & \textcolor{blue}{0.564} \\
  Ours & \textcolor{blue}{2.662} & 0.571 & \textcolor{red}{3.128} & \textcolor{blue}{0.556} & \textcolor{blue}{2.850} & 0.554 \\
  \Xhline{1pt} 
  \end{tabular}
\end{table*}

\begin{figure*}[t]
  \centering
   \includegraphics[width=0.95\linewidth]{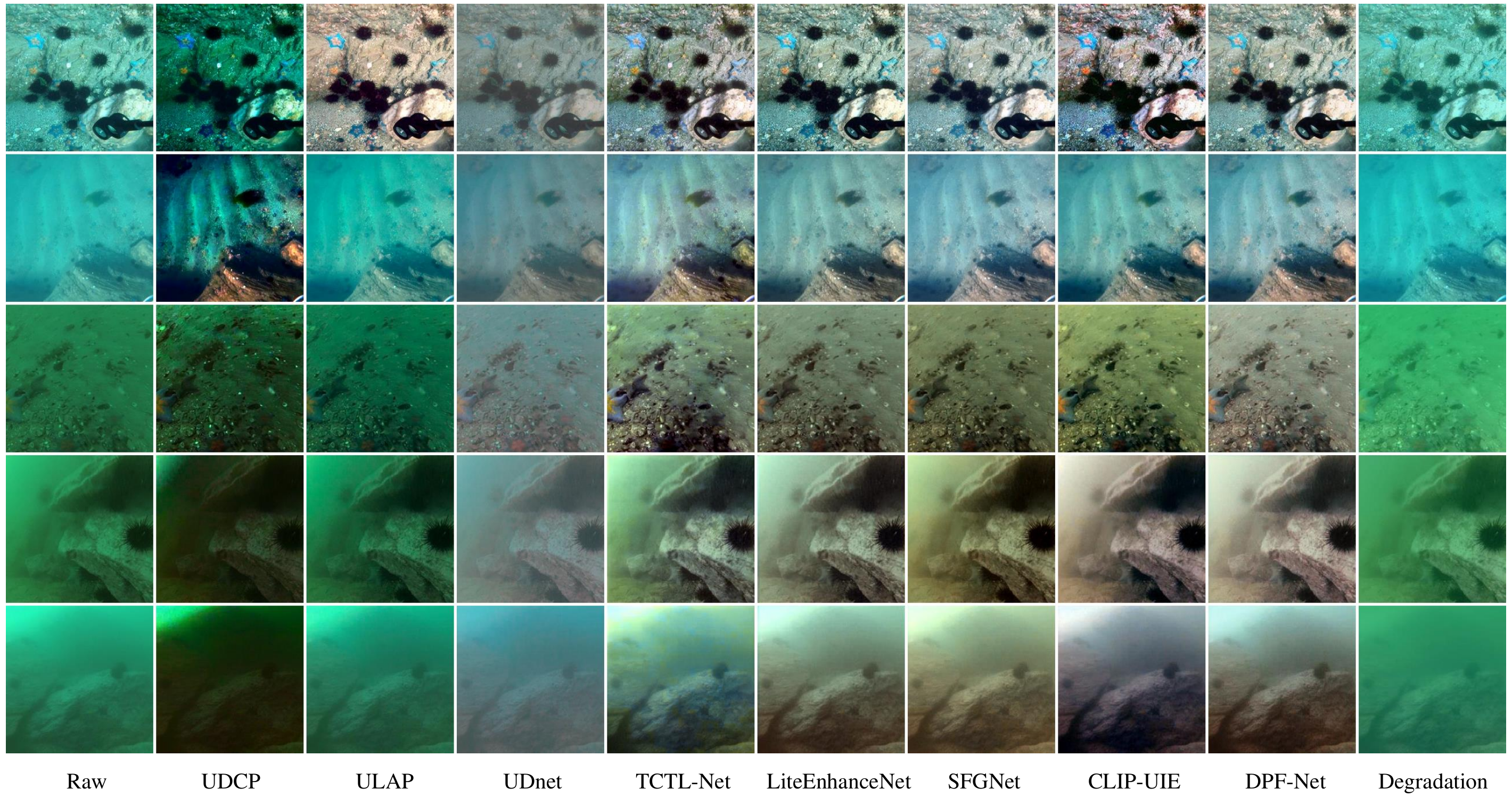}
   \caption{Visual comparisons on the images from the RUIE-ABC60 and RUIE-DE40 test set, which has no paired reference images. Each row, from top to bottom, corresponds to an image from subsets A through E, respectively. From left to right, raw underwater images and the results of UDCP \cite{drews2016underwater}, ULAP \cite{song2018rapid}, UDnet \cite{saleh2022adaptive}, TCTL-Net \cite{li2023tctl}, LiteEnhanceNet \cite{zhang2024liteenhancenet}, SFGNet \cite{zhao2024toward}, CLIP-UIE \cite{liu2024underwater}, the proposed DPF-Net and the predicted degradation images are presented, respectively.}
   \label{fig:RUIE_result}
\end{figure*}

Furthermore, we have conducted a comprehensive statistical analysis of the distribution of the RGB channels within the test set. As illustrated in Figure \ref{fig:UIEB&SUIM_result}, the bottom row depicts the distribution of the mean values of the RGB channels for all images from the UIEB-T dataset. The overall distribution of the raw images reveals a notably lower mean value in the red channel compared to the green and blue channels. In contrast, the mean distribution of the RGB channels in the enhanced images produced by DPF-Net demonstrates a more concentrated and balanced distribution among the three channels compared to other methods, indicating its superior capability in color correction and compensation. Furthermore, the distribution of the mean values in our predicted degradation images exhibits a high degree of consistency with that of the raw images, which aligns with the characteristics of underwater imaging.

\begin{figure}[t]
  \centering
   \includegraphics[width=1\linewidth]{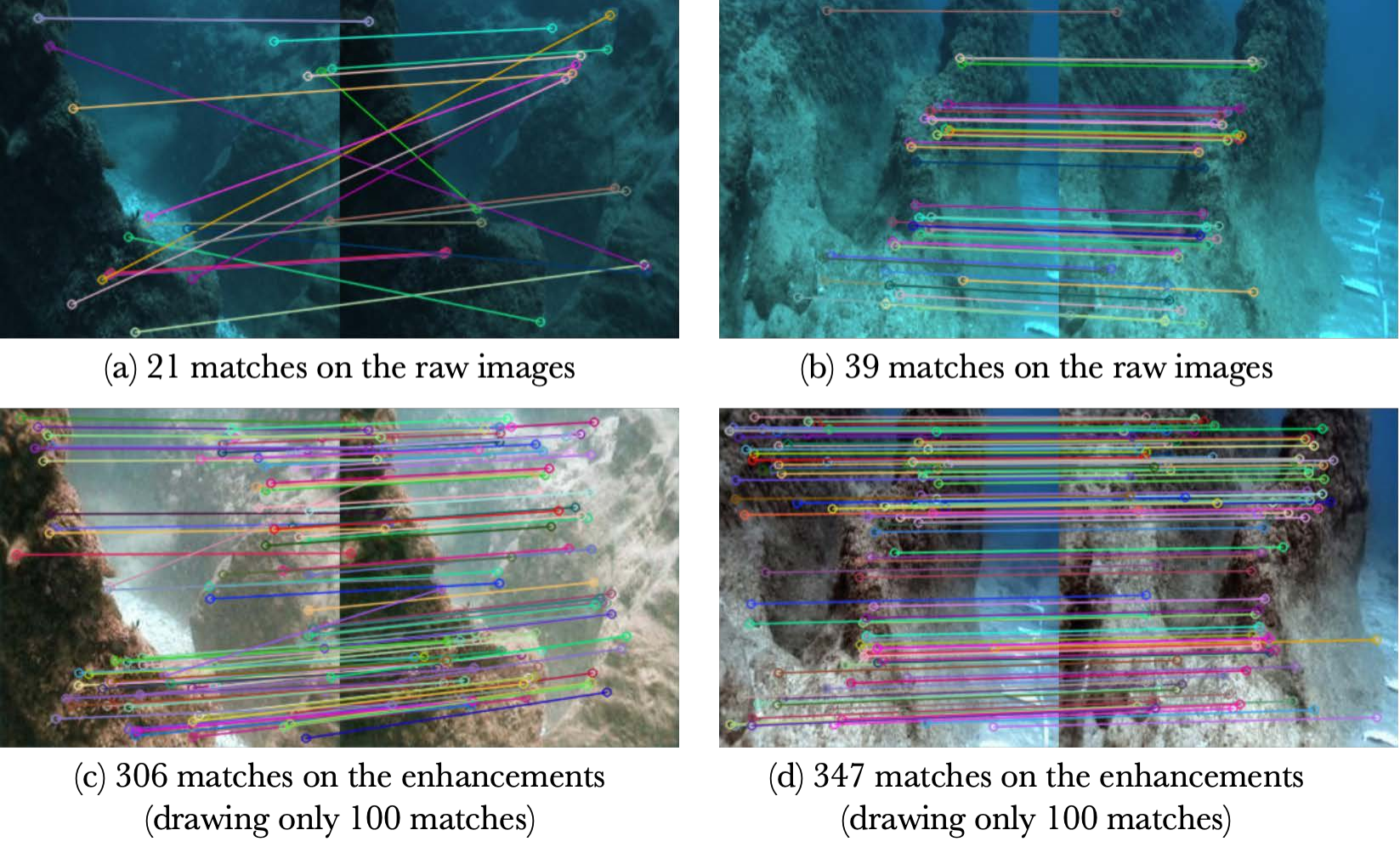}
   \caption{Matching of degraded raw underwater images ((a) and (b)) and enhanced underwater images ((c) and (d)). With the enhancement of DPF-Net, more matching can be achieved.}
   \label{fig:SIFT}
\end{figure}

As to the datasets with no reference images, the test images from UIEB-Challenging, RUIE-ABC60 and RUIE-DE40, despite possessing a predominantly uniform hue (predominantly blue and green), present entirely various imaging conditions for the networks and typically demonstrate significant degradation. The quantitative evaluations of the compared methods are shown in Table \ref{tab:C60&RUIE}. Figure \ref{fig:C60_result} (UIEB-Challenging) and Figure \ref{fig:RUIE_result} (RUIE-ABC60 and RUIE-DE40) illustrate the examples of the raw images, the enhanced images generated by each method, and the predicted degradation images. On these test sets, DPF-Net achieves superior enhancement performance, ranking within the top 3 across all six evaluation metrics and securing a top 2 position in four of them. These results effectively demonstrate the robust generalization capability of the proposed method. While CLIP-UIE achieves satisfactory visual effects on certain images, it produces inconsistent enhancement outcomes due to the inherent stochastic nature of the diffusion model, resulting in variability in output quality. In contrast, the physical model employed by DPF-Net demonstrates notable robustness, and the $L_{Lab}$ metric ensures consistent color improvements in enhanced images.

The significance of underwater image enhancement for subsequent tasks cannot be overstated. Taking the Scale-Invariant Feature Transform (SIFT) \cite{lowe2004distinctive}, an algorithm for keypoint matching, as an example, enhanced images enable the detection of a significantly greater number of keypoints compared to raw images. As illustrated in Figure \ref{fig:SIFT}, when applying SIFT to a raw image from the FLsea dataset \cite{randall2023flsea}, which comprises multiple video frame sequences and camera-sensor orientations, only a limited number of matching points are identified, with some being incorrectly matched. In contrast, the enhanced images yield a substantially higher number of accurate matching points. Experimental results demonstrate that the enhanced images generated by DPF-Net perform notably better in subsequent tasks.

\begin{table}[t]
  \centering\small
  \caption{Evaluation results of the ablation for DPEM on UIEB-T, verifying the effectiveness of DPEM}
  \fontsize{9pt}{12pt}\selectfont
  \label{tab:ablation_DPEM}
   \begin{tabular}{C{0.25\linewidth} | C{0.1\linewidth}  C{0.1\linewidth} C{0.1\linewidth}  C{0.1\linewidth} }
  \Xhline{1pt} 
    Datasets &  \multicolumn{4}{c}{UIEB-T}\\
  \hline
  Methods/Metrics    & PSNR$\uparrow$  & SSIM$\uparrow$ & UIQM$\uparrow$  & UCIQE$\uparrow$ \\
  \hline
  BL & 24.187 & 0.792 & 2.911 & 0.590\\
  BL+D & 24.678 & 0.808 & 2.991 & 0.601\\
  Full Model & 25.358 & 0.810 & 3.024 & 0.605\\
  \Xhline{1pt} 
  \end{tabular}
\end{table}

\begin{figure}[t]
  \centering
   \includegraphics[width=1\linewidth]{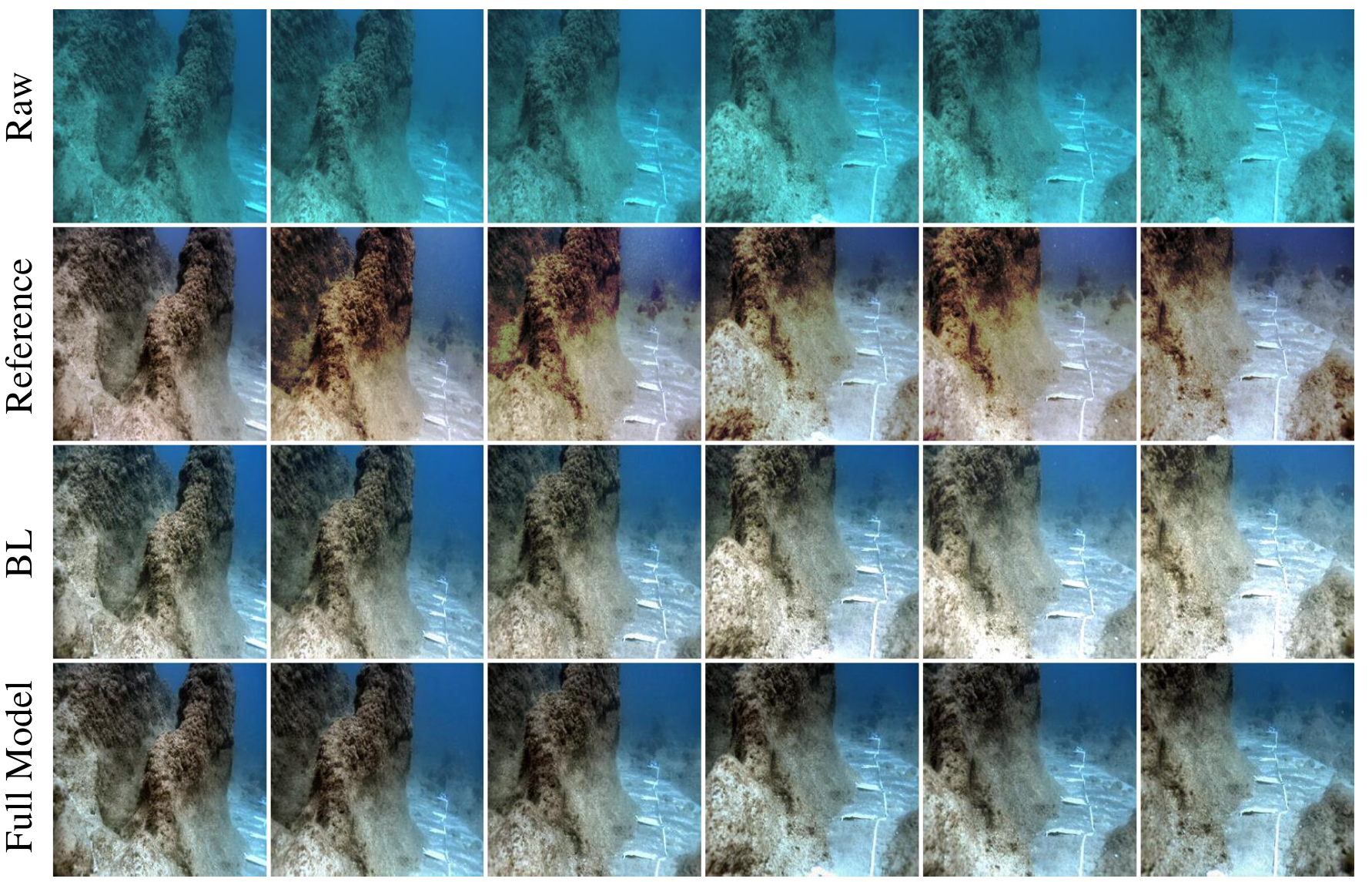}
   \caption{Each row, from top to bottom, sequentially represents the raw image, the reference image, the results obtained using BL (without DPEM), and the results generated by the full DPF-Net, respectively. The frame order is from left to right.}
   \label{fig:frames}
\end{figure}

\subsection{Ablation Study and Analysis}
To demonstrate the effectiveness of the key components of our proposed framework, we conducted a series of ablation studies on the UIEB training set, adhering to the consistent training methodology described previously. The results were evaluated using PSNR, SSIM, UIQM, and UCIQE metrics. The ablation study is primarily divided into three parts: validating the effectiveness of DPEM, examining the training strategy for DPEM, and evaluating $L_{Lab}$ independently. The details of each test model are as follows:
\begin{itemize}
    \item \textbf{Full Model}: The complete DPF-Net framework introduced in this study.
    \item \textbf{BL} (BaseLine): The hybrid architecture, which integrates CT-UNet and VQ-VAE, serves as the backbone of the enhancement network and is trained using $L_{ref}$.
    \item \textbf{BL+D}: Incorporating DPEM to integrate physical parameters into the training process of the BL model.
    \item \textbf{BL+L}: Adding the weakly supervised reference loss $L_{Lab}$ into the training process of the BL model.
    \item \textbf{BL+D+L}: Incorporating DPEM alongside $L_{Lab}$ into the BL model, while omitting the degradation consistency loss $L_{deg}$.
\end{itemize}

\textbf{Ablation for DPEM.} DPEM's function is to supply estimated physical parameters, integrate them into the VQ-VAE embedding space for feature extraction, and generate degradation images based on these parameters for the enhanced image obtained by the network, thereby facilitating the computation of degradation consistency loss. We trained the BL, BL+D, and Full Model using identical experimental configurations. The evaluation results presented in Table \ref{tab:ablation_DPEM} show the performance improvement of incorporating physical knowledge. Moreover, throughout the training phase, we noted that the loss of BL exhibited minor oscillations midway, failing to reduce or converge efficiently. Nonetheless, following the introduction of D, the loss of BL+D exhibited markedly enhanced stability throughout training and was able to converge efficiently. Besides, the visual improvement impact of the model achieved using BL+D in the test set exhibited greater stability than BL, avoiding both excessive and insufficient augmentation. We assert that the implementation of DPEM can effectively leverage the resilience of physical knowledge to impose additional constraints on model training, moving it beyond a purely data-driven approach.

\begin{figure}[t]
  \centering
   \includegraphics[width=0.95\linewidth]{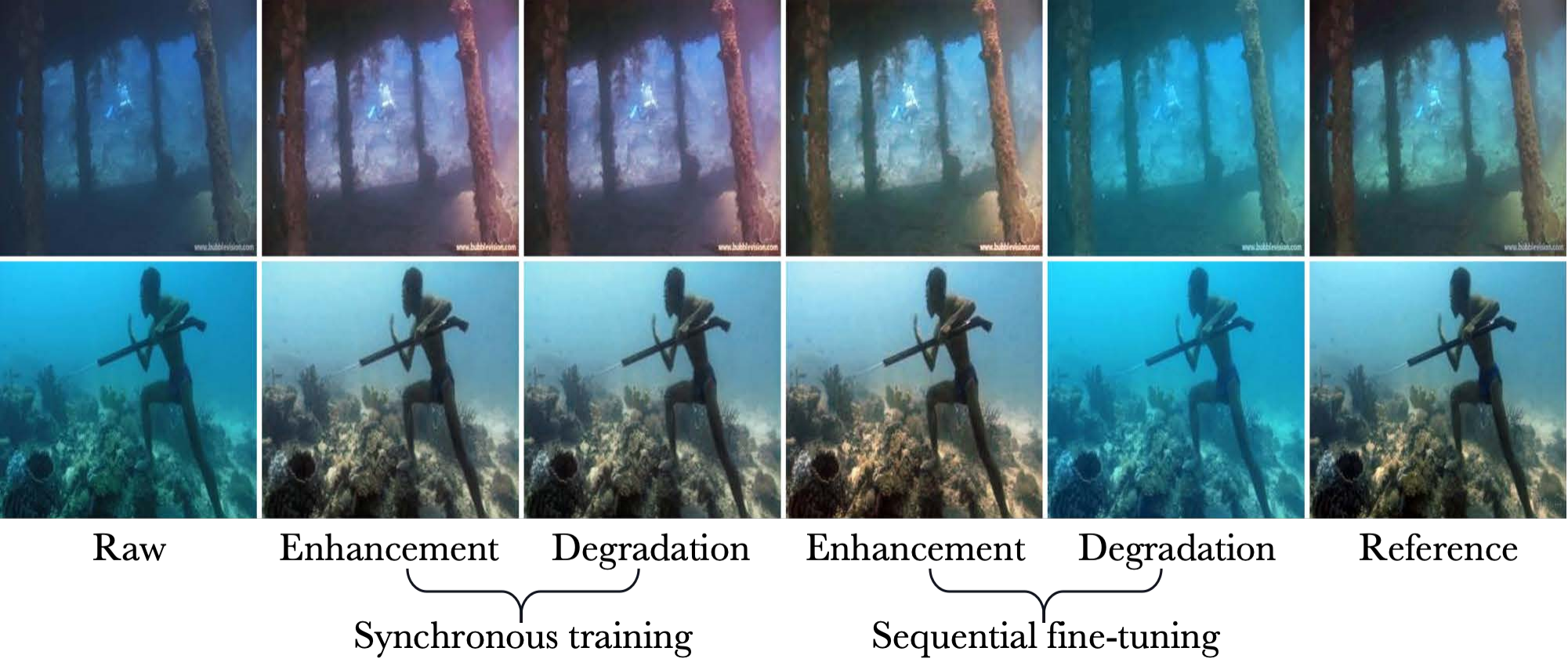}
   \caption{Each column, from left to right are the raw, enhanced and predicted degraded images of synchronous training, enhanced and predicted degraded images of sequential fine-tuning and reference respectively. The inaccuracy of the physical parameter estimation of synchronous training causes the predicted degraded images to be very different from the raw images.}
   \label{fig:training_strategy}
\end{figure}

\begin{table}[t]
  \centering\small
  \caption{Enhancement performance evaluation of the ablation on training strategies. ``synchronous training" means that the DPEM and DPF-Net are trained together from scratch, and ``Sequential fine-tuning" (the aforementioned Full Model) means that the DPEM is trained on the synthetic dataset first and then fine-tuned with DPF-Net.}
  \fontsize{9pt}{12pt}\selectfont
  \label{tab:ablation_strategy}
   \begin{tabular}{C{0.35\linewidth} | C{0.1\linewidth}  C{0.1\linewidth} C{0.1\linewidth}  C{0.1\linewidth} }
  \Xhline{1pt} 
    Datasets &  \multicolumn{4}{c}{UIEB-T}\\
  \hline
  Methods/Metrics    & PSNR$\uparrow$  & SSIM$\uparrow$ & UIQM$\uparrow$  & UCIQE$\uparrow$ \\
  \hline
  Synchronous training & 23.164 & 0.740 & 2.892 & 0.600\\
  Sequential fine-tuning & 25.358 & 0.810 & 3.024 & 0.605\\
  \Xhline{1pt} 
  \end{tabular}
\end{table}

\begin{table}[t]
  \centering\small
  \caption{Parameter prediction performance evaluation of the ablation on training strategies. SSIM and PSNR metrics are employed to measure the similarity between the predicted degradation images, generated using the predicted parameters of DPEMs trained with the two training strategies, and the original underwater images.} 
  \fontsize{9pt}{12pt}\selectfont
  \label{tab:ablation_strategy_para}
   \begin{tabular}{C{0.35\linewidth} | C{0.2\linewidth}  C{0.2\linewidth}}
  \Xhline{1pt} 
    Datasets &  \multicolumn{2}{c}{UIEB-T}\\
  \hline
  Methods/Metrics    & PSNR$\uparrow$  & SSIM$\uparrow$ \\
  \hline
  Synchronous training & 18.961 & 0.809 \\
  Sequential fine-tuning & 20.974 & 0.849 \\
  \Xhline{1pt} 
  \end{tabular}
\end{table}

We also observed that DPEM exhibits a stabilizing effect on the enhancement of consecutive underwater frames, thereby preserving visual continuity between them. Since the physical parameters of inter-frame images generally remain consistent, DPF-Net can ensure output image stability by employing nearly identical physical parameters. We present a comparison of a series of enhanced consecutive frames from the FLsea dataset in Figure \ref{fig:frames}. The reference image has significant hue distortion. BL (without DPEM) shows an obvious phenomenon of luminous instability in the enhanced consecutive frames. Conversely, the complete DPF-Net model effectively demonstrates the stability imparted by the physical model, achieving overall commendable visual coherence.

\textbf{Ablation for training strategy of DPEM.} To further validate the efficacy and necessity of our two-stage training strategy in our data-driven approach supported by the physical imaging model, we re-trained the DPEM and DPF-Net using a synchronous training strategy. That is, instead of employing a pre-trained DPEM to forecast physical parameters, we concurrently train both DPEM and DPF-Net from scratch. The loss function of DPF-Net remained unchanged; however, DPEM no longer has actual physical parameters for supervision, and its supervision is only from $L_{deg}$. The enhancement performance on UIEB-T of the two DPF-Net trained using these two different training strategies, i.e., synchronous training and sequential fine-tuning (see Section \ref{sec:details}), is presented in Table \ref{tab:ablation_strategy}. Their visual comparisons of enhancements are illustrated in Figure \ref{fig:training_strategy}. It is evident that the use of a synchronous training strategy led to a decline in both visual quality and evaluation metrics of the enhanced images.

This indicates that DPEM cannot have a significant positive effect on the image enhancement task during synchronous training. We also observed an obvious decrease in the prediction performance of the parameters, evident in certain instances where the degraded images generated by DPF-Net displayed significant discrepancies compared to the original images. To quantitatively assess the prediction performance of parameters for two DPEMs using the two training strategies on natural underwater images, we evaluate the similarity between the predicted degradation images, generated using their respective predicted parameters, and the original underwater images. We used SSIM and PSNR metrics for this evaluation, with the results on UIEB-T presented in Table \ref{tab:ablation_strategy_para}. From the quantitative results, it can be clearly seen that sequential fine-tuning can enable DPEM to achieve better prediction performance for degraded parameters. Theoretically speaking, what synchronous training acquires is merely a ``fitting" imaging model from raw images to references, rather than the reliable degradation model.

\begin{figure}[t]
  \centering
   \includegraphics[width=0.9\linewidth]{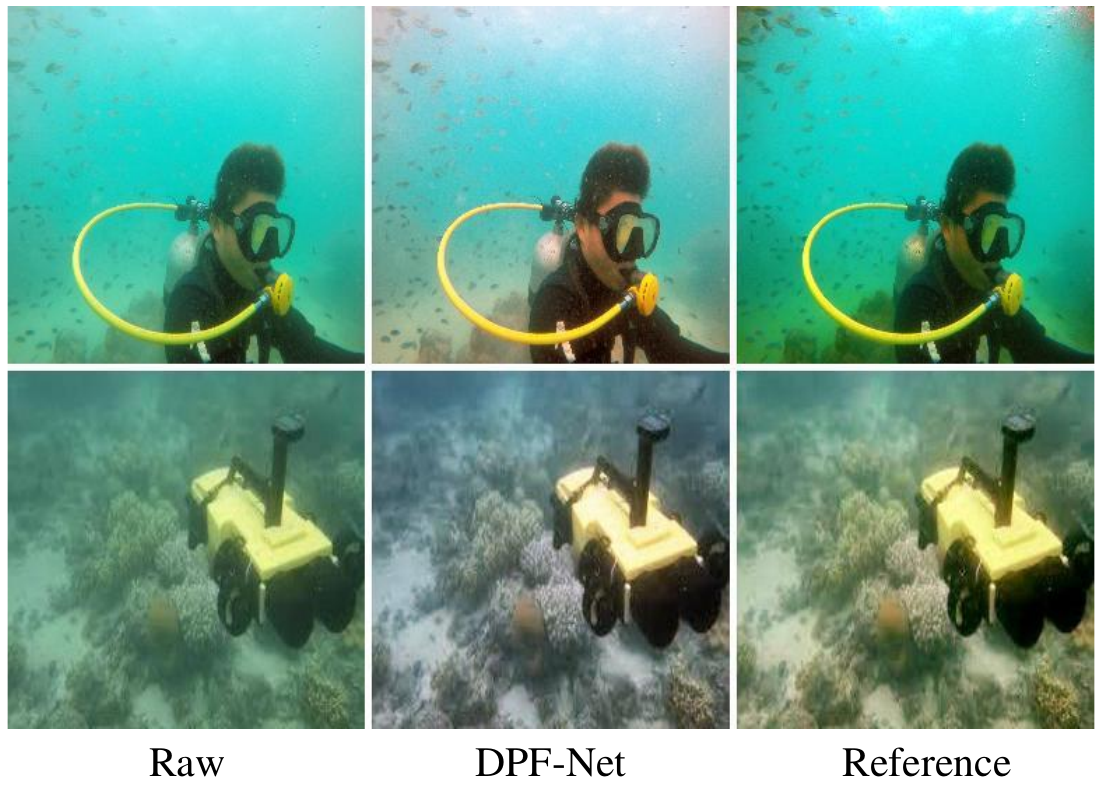}
   \caption{Each row, from left to right are the raw images from the training set, enhanced output during the training, and reference images respectively. It's noticeable that during training, the model's outputs are not overly influenced by the low-quality reference image.}
   \label{fig:training_img}
\end{figure}

\textbf{Ablation for $L_{Lab}$.} Our design concept for $L_{Lab}$ aims to have the enhanced image in the Lab color space approximate the distribution of the entire reference collection, rather than being limited to the quality of a single paired reference image. Similar to how color transfer methods \cite{reinhard2001color} alter the color scheme of an image, $L_{Lab}$ can guide the color distribution of the enhanced image to be more reasonable. We trained the BL model, the BL+L model, and the Full Model within an identical experimental framework. Throughout the training process, $L_{Lab}$ exhibited stable convergence and did not substantially influence the overall loss trajectory of the network. As anticipated, the hues of the enhanced images became more vivid, which was also evident in the enhancement of no-reference evaluation measures. As demonstrated in Figure \ref{fig:training_img}, our network produced more plausible enhancements on the training set images compared to the provided references, even when low quality reference images were utilized. The evaluation results of the enhancement performance are shown in Table \ref{tab:ablation_L}, which further demonstrates that the implementation of this weakly supervised loss diminished the influence of substandard reference images on the network.

\begin{table}[t]
  \centering\small
  \caption{Evaluation results of the ablation for $L_{Lab}$ on UIEB-T, verifying the effectiveness of $L_{Lab}$}
  \fontsize{9pt}{12pt}\selectfont
  \label{tab:ablation_L}
   \begin{tabular}{C{0.25\linewidth} | C{0.1\linewidth}  C{0.1\linewidth} C{0.1\linewidth}  C{0.1\linewidth} }
  \Xhline{1pt} 
    Datasets &  \multicolumn{4}{c}{UIEB-T}\\
  \hline
  Methods/Metrics    & PSNR$\uparrow$  & SSIM$\uparrow$ & UIQM$\uparrow$  & UCIQE$\uparrow$ \\
  \hline
  BL & 24.187 & 0.792 & 2.911 & 0.590\\
  BL+L & 24.306 & 0.805 & 2.960 & 0.602\\
  Full Model & 25.358 & 0.810 & 3.024 & 0.605\\
  \Xhline{1pt} 
  \end{tabular}
\end{table}

\section{Discussion and future works}\label{sec:Discussion}
It should be known that the physical parameters assessed by DPEM may not be entirely accurate due to the inherent complexity and variability of underwater environments. Although the absolute depth scale and background light can be visually evaluated for plausibility, the attenuation and scattering coefficients cannot be reliably quantified. Consequently, the physical parameters primarily serve a qualitative function, rather than providing precise quantitative measures within our network design. Currently, we have leveraged the strengths of the physical imaging model, which offers numerous unexplored advantages. In the future, we will concentrate on developing quantitative evaluation methods for the physical parameters of actual underwater images to explore the potential synergy between data-driven deep learning and physical imaging models. Additionally, we will further investigate the integration of physical parameters with other tasks beyond UIE.

\section{Conclusion}\label{sec:Conclusion}
This paper introduces a novel network, termed the Data-Driven and Physical Parameters Fusion Network (DPF-Net), designed to enhance underwater images. By integrating the physical model of underwater imaging with deep learning methodologies, DPF-Net aims to leverage the strengths of both approaches to achieve optimal performance. In our two-stage framework, we initially train DPEM using a synthetic dataset. Subsequently, we integrate the pre-trained DPEM with the enhancement network DPF-Net through a combined approach that incorporates space embedding and a weighted fusion strategy for joint training. Furthermore, we present a series of loss functions for training, comprising supervision loss between the raw and reference images, degradation consistency loss, and a novel weak supervised reference loss derived from the Lab distribution across the complete reference set. In comparable experiments, we achieved state-of-the-art results that showcasing substantial advantages in both visual effects and evaluation metrics. Our proposed framework aims to integrate an accurate physical imaging model with deep learning methodologies. We will also explore the potential synergies between these two approaches in our subsequent research.

\ifCLASSOPTIONcaptionsoff
  \newpage
\fi

\bibliographystyle{IEEEtran}

\bibliography{reference}


\end{document}